\documentclass[11pt]{article}

\usepackage{amsmath, amssymb}
\usepackage{graphicx}
\usepackage{hyperref}
\usepackage[numbers,sort&compress]{natbib}
\usepackage{geometry}
\usepackage{times}
\usepackage{float}
\usepackage{placeins}
\usepackage{flafter}
\usepackage{multirow}
\usepackage{booktabs}
\usepackage{caption}
\usepackage{subcaption}
\usepackage[table]{xcolor}
\usepackage{xspace}
\usepackage{enumitem}
\definecolor{linkblue}{HTML}{1C4E91}
\hypersetup{colorlinks=true, linkcolor=linkblue, citecolor=linkblue, urlcolor=linkblue}
\usepackage{pgfplots}
\pgfplotsset{compat=1.18}
\definecolor{cOurs}{HTML}{2E8B57}
\definecolor{cSim}{HTML}{DC143C}
\definecolor{cHiLo}{HTML}{F2C200}
\definecolor{cHiLoEdge}{HTML}{9A7B00}
\definecolor{cAlt}{HTML}{4169E1}
\definecolor{cSalmon}{HTML}{FF8C8C}
\definecolor{cSalmonLight}{HTML}{FFC4C4}
\definecolor{cBase}{HTML}{BDBDBD}
\definecolor{cRef}{HTML}{8C8C8C}
\definecolor{cGrid}{HTML}{DADADA}
\pgfplotsset{paper/.style={
  font=\footnotesize, axis lines=box, axis line style={black!85, line width=0.6pt},
  tick style={black!85, line width=0.5pt}, tick align=outside, xtick pos=bottom, ytick pos=left,
  grid=major, grid style={cGrid, line width=0.4pt},
  legend style={draw=black!25, rounded corners=2pt, fill=white, fill opacity=0.9, text opacity=1,
    font=\footnotesize, inner xsep=3pt, inner ysep=2pt, row sep=-1pt},
  legend cell align=left,
  every axis plot/.append style={line width=1.4pt, mark size=2.2pt}}}

\newcommand{\method}{Robust-GCD\xspace}
\newcommand{\setting}{D-GCD\xspace}
\newcommand{\cifarten}{\textit{CIFAR-10}\xspace}
\newcommand{\cifarhundred}{\textit{CIFAR-100}\xspace}
\newcommand{\cub}{\textit{CUB}\xspace}
\newcommand{\cars}{\textit{Stanford Cars}\xspace}
\newcommand{\aircraft}{\textit{FGVC-Aircraft}\xspace}

\newcommand{\Dl}{\mathcal{D}^{l}}
\newcommand{\Du}{\mathcal{D}^{u}}

\newcommand{\Yk}{\mathcal{Y}_{k}}
\newcommand{\Yn}{\mathcal{Y}_{n}}
\newcommand{\corrupt}[2]{\mathrm{corrupt}_{#1,#2}}
\newcommand{\sg}{\mathrm{sg}}

\title{Robust Generalized Category Discovery}
\author{
  Daniel Wang \\
  \normalsize Wellesley High School \\
  \normalsize danielxwang.27@gmail.com \\
}
\date{\today}

\begin{document}

\maketitle
\begin{abstract}
Generalized Category Discovery (GCD) classifies an unlabeled pool that contains both known and novel classes, given labels for the known classes only. Existing GCD methods assume that every unlabeled image is clean, but unlabeled data collected in the wild is often degraded by blur, sensor noise and compression artifacts. We formalize the setting in which an unknown fraction of the unlabeled pool is corrupted by an unknown operator as Degraded Generalized Category Discovery (\setting). To tackle it, we propose \method, a framework that estimates the corruption directly from the unlabeled pool and uses the estimate to build labeled anchors for known classes and pseudo-anchors for novel classes inside the corrupted region of feature space. A trust curriculum controls how much the unlabeled pool is trusted while the anchors take hold, and a paired alignment loss keeps a clean image and its corrupted twin close in feature space. Across five datasets and multiple corruption families, \method recovers between 43\% and 96\% of the All accuracy that SimGCD loses under contamination, and more than 60\% on four of the five datasets, with the largest gains on the corrupted images themselves.
\end{abstract}

\section{Introduction}
\label{sec:intro}

Generalized Category Discovery (GCD)~\citep{vaze2022gcd} asks a model to classify an unlabeled pool of images that contains both known and novel classes, given labeled images of the known classes only. It extends novel class discovery~\citep{han2019dtc}, which assumes that every unlabeled image belongs to a novel class. Most GCD methods build on self-supervised vision transformers~\citep{caron2021dino}, such as SimGCD~\citep{wen2023simgcd} and its successors~\citep{vaze2023norep, wang2024sptnet, choi2024cms}. These methods are developed and evaluated on benchmarks~\citep{vaze2022ssb} in which every image of the unlabeled pool is clean.

Real unlabeled pools, however, are rarely clean. Images gathered from cameras, phones and the web are often degraded by blur, sensor noise, compression and similar corruptions~\citep{hendrycks2019benchmarking}, which are known to affect deep networks~\citep{dodge2016quality, geirhos2018generalisation}. In GCD this corruption falls mainly on the unlabeled pool, because the labeled set is curated by hand. There are many different techniques to learn from corrupted data, but each assumes something that GCD does not provide. Augmentation-based training~\citep{hendrycks2020augmix}, invariance learning~\citep{arjovsky2019irm} and test-time adaptation~\citep{wang2021tent} assume that every class has labeled examples, while the novel classes in GCD have none. Adversarial training~\citep{madry2018pgd} requires the perturbation set to be fixed in advance, while the corruption in the pool is unknown. Label-noise methods~\citep{li2020dividemix} assume that the noise lies in the labels, while here it lies in the images. The closest prior work in Generalized Category Discovery is domain-shift GCD~\citep{wang2025hilo, feng2025free}, which trains on an unlabeled pool that contains images from a shifted domain. It does not fit a corrupted pool for two reasons. (a) It treats the shifted images as a domain whose information is separated from semantic information, while a corruption is a nuisance that damages individual images of any class. (b) It does not estimate the operator that produced the shifted images, so it cannot apply it to labeled images to supervise the corrupted region of feature space.

We therefore study Degraded Generalized Category Discovery (\setting), a GCD setting in which an unknown fraction of the unlabeled pool has been corrupted by an unknown operator. As in GCD, the model is given a clean labeled set of known classes and an unlabeled pool of known and novel classes, and it must assign every pool image to a known or novel class. Unlike GCD, the corrupted images are unmarked and mixed with clean ones, and the operator damages their appearance without changing their class. Unlike domain-shift GCD, the corruption is a nuisance to estimate from the pool, not a domain to learn. This raises two challenges. First, the corruption is unobserved, since the model knows neither which images are corrupted nor which operator corrupted them. Second, novel classes have no labels, so they are organized only by similarity among unlabeled images, which the corrupted images make unreliable.

To address these challenges, we propose \method, which estimates the corruption from the unlabeled pool and uses it to supervise corrupted images. An estimation stage finds the suspected corrupted part of the pool and matches it to a library of synthetic corruptions~\citep{michaelis2019imagecorruptions}, without corruption labels. A training stage applies the matched corruption to labeled images to create anchors for known classes, and to confident predictions on unlabeled images to create pseudo-anchors for novel classes. A paired alignment loss and a trust curriculum protect the features while the anchors take hold. Our contributions are as follows.

\begin{itemize}[wide=10pt, leftmargin=*, nosep]
    \item We formalize \setting, a GCD setting in which an unknown fraction of the unlabeled pool is corrupted by an unknown operator, which unifies category discovery and learning from corrupted data, and we identify challenges that existing techniques in both fields do not address.
    \item We propose \method, which estimates the corruption from the unlabeled pool without corruption labels and uses the estimate to extend supervision to corrupted images of both known and novel classes.
    \item We construct a \setting benchmark protocol on five generic and fine-grained datasets with multiple corruption families. \method improves All accuracy over SimGCD on all five datasets and recovers 43\% to 96\% of the All accuracy that SimGCD loses to contamination.
\end{itemize}

\section{Related Work}
\label{sec:related}
We review three lines of work: Generalized Category Discovery, Domain-Shift Generalized Category Discovery, and Robust Representation Learning.

\textbf{Generalized Category Discovery} (GCD)~\citep{vaze2022gcd} is an open-world machine learning task to classify an unlabeled pool containing both known and novel classes, given a labeled set of known classes only. The original method in the setting fine-tunes a Vision Transformer (ViT) pretrained without labels by self-distillation (DINO)~\citep{caron2021dino}
with contrastive objectives and obtains predictions with semi-supervised $k$-means. SimGCD, a state-of-the-art baseline GCD method, directly creates a classifier by using a self-distillation objective, which forces augmented views of image to match, and uses a mean-entropy regularizer to force relatively balanced predictions across all categories ~\citep{wen2023simgcd}. More recent work considers harder settings, including long-tailed training sets~\citep{bai2023distagnostic} and continual discovery without forgetting~\citep{ma2024happy}. Nonetheless, all of these methods assume that every image in the unlabeled pool is well-formed. To our knowledge, no prior GCD work studies the scenario where the unlabeled pool contains an unknown fraction of corrupted images. 

\textbf{Domain-Shift Generalized Category Discovery} (DS-GCD) was proposed by HiLo~\citep{wang2025hilo}, which takes a domain feature from an early block of a ViT and a semantic feature from the last block, minimizes the mutual information between them by training a small critic network to distinguish feature pairs, and admits shifted images into training gradually through a curriculum. FREE~\citep{feng2025free} separates images from the known and unknown domains using their frequency spectra, and swaps low-frequency amplitude between domains so that features become insensitive to the domain. DS-GCD is the closest setting to ours, using similar benchmark families. In DS-GCD, however, the shifted images represent a domain which the model learns to separate from semantic content, while in \setting, the corruption is an unknown percent whose operator must be estimated from the pool before it can be used.

\textbf{Robust Representation Learning} covers techniques to learn from corrupted inputs, branching into four families of methods.
Adversarial robustness~\citep{madry2018pgd} trains against worst-case
perturbations found by an optimizer. Domain generalization and
out-of-distribution methods~\citep{arjovsky2019irm} remove
environment-specific shortcuts so that the retained features transfer across
conditions. Self-supervised invariance~\citep{chen2020simclr} constructs its own
targets by requiring two augmented views of an image to map to the same
representation. Label-noise and latent-filtering
methods~\citep{li2020dividemix} track per-sample loss to separate reliable
training examples from unreliable ones.

In summary, most prior GCD work assumes a clean pool, while domain-shift GCD treats shifted images as a domain to be separated from semantics, and robust representation learning assumes labeled examples for every class, noise in the labels, or a perturbation known in advance. \setting fails each of these, since its corruption is an unknown nuisance on unlabeled images of both known and novel classes.

\section{Problem Formulation and Motivation}
\label{sec:prelim}

\subsection{Motivation}
\label{sec:motivation}

In practice, the unlabeled pool is the part of a GCD dataset most likely to contain corrupted images. Unlabeled images are collected at scale from cameras, phones, and other online resources, which are susceptible to many different corruptions, such as motion blur from camera shake, sensor noise in low light, or compression artifacts from repeated re-encoding~\citep{hendrycks2019benchmarking}. 
We expect contamination to negatively affect accuracy. Every class, known or novel, is represented by a prototype, a learnable vector in feature space, and each image is assigned to the prototype it is most similar to. Each known class has a set of clean labeled images, and their labels pull its prototype toward those images. A novel prototype, in contrast, is placed only by similarity among unlabeled images, and some of those images are corrupted. A corruption changes the appearance of an image without changing its class, and deep networks are known to be sensitive to such changes~\citep{dodge2016quality, geirhos2018generalisation}. If a corruption moves an image's features further than its class does, corrupted images will be more similar to one another than to clean images of their own class.

\subsection{Problem Formulation}
\label{sec:formulation}

Let $\mathcal{X}$ denote the image space and $\mathcal{Y} = \Yk \cup \Yn$ the
label space, where $\Yk$ is the set of known classes, $\Yn$ is the set of novel
classes, and $\Yk \cap \Yn = \emptyset$. The training set consists of a labeled
set $\Dl = \{(\boldsymbol{x}^l_i, y^l_i)\}_{i=1}^{N_l} \subset \mathcal{X} \times \Yk$ where $\boldsymbol{x}^l_i$ is the $i$-th labeled image, $y^l_i \in \Yk$ is its class, and $N_l$ is the number of labeled images. The training set also consists of an unlabeled pool
$\Du = \{\boldsymbol{x}^u_i\}_{i=1}^{N_u} \subset \mathcal{X}$, where $\boldsymbol{x}^u_i$ is the $i$-th unlabeled image and $N_u$ is the number of unlabeled images. Each $\boldsymbol{x}^u_i$ belongs to a class in $\Yk \cup \Yn$ that the model does not observe. The composition of $\Du$ contains a
clean part $\Du_{\text{cln}}$ and a corrupted part $\Du_{\text{cor}}$, such that
\begin{equation}
\label{eq:rho}
  \Du = \Du_{\text{cln}} \cup \Du_{\text{cor}},
  \qquad
  \rho = \frac{|\Du_{\text{cor}}|}{|\Du|}.
\end{equation}
We refer to $\rho$ as the contamination
rate. Each image in $\Du_{\text{cor}}$ has been altered by a corruption, an image operator such as blur or noise that damages appearance but does not change the class. We call this setting Degraded Generalized Category Discovery (\setting). The labeled
set $\Dl$ is clean, which reflects how labeled data is manually
curated in practice. In \setting, the corrupted images are part of the training pool, mixed with clean images and unmarked, while no knowledge of $\rho$, which images belong to $\Du_{\text{cor}}$, or the operator that produced them are known.

The objective of \setting is to assign every image in
$\Du$ to a class in $\mathcal{Y}$, recognizing the known classes and discovering novel ones. Following common practice~\citep{vaze2022gcd, wen2023simgcd} we
treat the total class count $|\mathcal{Y}|$ as a known prior. We report
clustering accuracy after matching clusters to classes through Hungarian assignment on $\Yk$ (Known), $\Yn$ (Novel), and
$\Yk \cup \Yn$ (All), and split each by whether the image is clean
or corrupted.

\subsection{Revisiting SimGCD}
\label{sec:simgcd}

Our relevant components from SimGCD for our method include a backbone $f(\cdot)$ which maps an image to a feature
${h}_i = f({x}_i)$ for the $i$-th image in a batch, and a set of prototypes $\{\boldsymbol{\mu}_k\}_{k=1}^{|\mathcal{Y}|}$. Each image is assigned to the prototype its feature is most similar to. The probability that image $i$ belongs to class $k$ is a softmax over cosine similarities with temperature $\tau_s$,
\begin{equation}
\label{eq:softmax}
  p_i^{(k)} = \frac{\exp\!\big(\tfrac{1}{\tau_s}\, {\cos(\boldsymbol{h}_i, \boldsymbol{\mu}_k)} \big)}
                   {\sum_{k'} \exp\!\big(\tfrac{1}{\tau_s}\, {\cos(\boldsymbol{h}_i, \boldsymbol{\mu}_{k'})} \big)},
\end{equation}
where $\cos(\cdot,\cdot)$ is cosine similarity and $\boldsymbol{p}_i$ is the vector of the probabilities $p_i^{(k)}$. Labeled images are trained with cross-entropy loss $\mathcal{L}_s$ against their ground-truth labels. All images are also trained with a self-distillation loss $\mathcal{L}_u$, in which each image is augmented twice, and the predictions on each view are trained to match each other with cross-entropy. This pushes the model to make consistent and confident predictions across augmentations.
A mean-entropy term is added on top to prevent the classifier from assigning all samples to a small number of
prototypes. $H(\cdot)$ represents Shannon entropy and $\bar{\boldsymbol{p}} =
\tfrac{1}{B}\sum_{i=1}^{B}\boldsymbol{p}_i$ for the class-probability
vector averaged over a batch of size $B$, the regularizer is
$-\varepsilon\, H(\bar{\boldsymbol{p}})$ with weight $\varepsilon$. We use $\mathcal{L}(\mathcal{D})$ to denote the loss averaged over samples drawn from $\mathcal{D}$. Omitting unchanged terms, the full loss for SimGCD is

\begin{equation}
\label{eq:simgcd}
  \mathcal{L}_{\text{SimGCD}}
  = \mathcal{L}_{s}(\Dl) + \mathcal{L}_{u}(\Dl \cup \Du) - \varepsilon\, H(\bar{\boldsymbol{p}}).
\end{equation}

Two properties of this objective are relevant. First, the supervised term is computed only on $\Dl$, so ground-truth
supervision reaches known-class prototypes and no others. Second, the entropy
term encourages the classifier to distribute predictions across prototypes, but
it does not constrain where a prototype is located in feature space.

\section{\method}
\label{sec:method}

\begin{figure}[!htb]
  \centering
  \includegraphics[width=\textwidth]{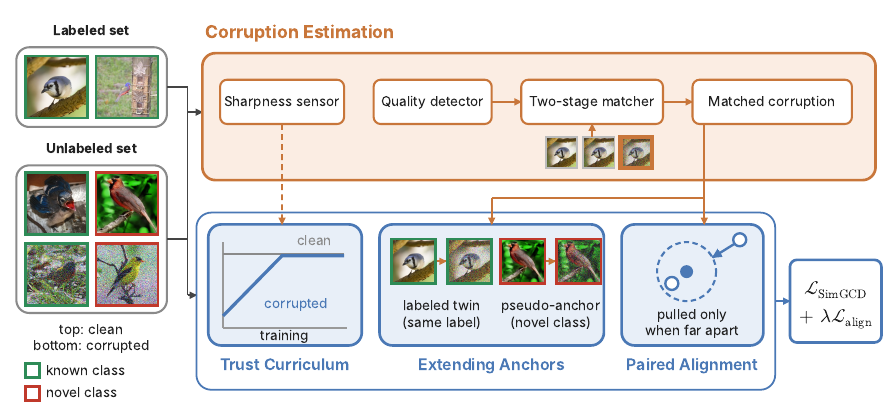}
  \caption{Overview of \method. The sensor scores unlabeled images for
  corruption (Section~\ref{sec:estimate}), and a separate quality detector flags the suspected corrupted
  part of the pool. The matcher converts that flagged set into a corruption
  operator $(c^\ast, s^\ast)$, which is applied to labeled images and, from
  epoch $E_{\text{pa}}$ onward, to novel-class pseudo-anchors (Section~\ref{sec:anchor}). A thresholded paired
  alignment loss ties each corrupted twin to its clean source, and a trust
  curriculum controls how strongly the sensor weights the unlabeled pool over
  training (Section~\ref{sec:align}).}
  \label{fig:framework}
\end{figure}

To tackle the \setting problem, we propose \method, which gives corrupted images the supervision that the novel prototypes lack. Section~\ref{sec:estimate} estimates the corruption from the unlabeled pool, Section~\ref{sec:anchor} uses the estimate to extend anchors into the corrupted region of feature space, and Section~\ref{sec:align} adds a trust curriculum that schedules how far the unlabeled pool is trusted and a paired alignment loss that keeps corrupted twins close to their clean sources. Figure~\ref{fig:framework} shows how these stages connect. 

\subsection{Corruption Estimation}
\label{sec:estimate}

This stage determines which unlabeled images are corrupted and which synthetic
operator produces comparable corruption on a clean image, using no corruption
labels.

\textbf{Corruption-aware sensor.} Most corruptions change the high-frequency
content of an image. For instance, blur removes it and most noise families add it. To detect this, our sensor assigns each unlabeled image a weight $w_i \in [0,1]$ that is low when the image is likely corrupted, using the variance of the Laplacian response, a second-derivative filter which measures how much high-frequency detail an image contains, 

\begin{equation}
\label{eq:sensor}
  v_i = \mathrm{Var}\big(\Delta \boldsymbol{x}_i\big),
  \qquad
  \mathrm{Var}(L) = \frac{1}{Q}\sum_{q=1}^{Q}\big(L_q - \bar{L}\big)^2,
\end{equation}
where $\Delta$ is the discrete Laplacian, $v_i$ is the sensor’s raw score for unlabeled image, $Q$ is the number of pixels, $L = \Delta\boldsymbol{x}_i$ is the Laplacian response map, $L_q$ is its value at pixel $q$, and $\bar{L}$ is the mean response. Since corruption can move $v_i$ in either
direction depending on the family, we standardize $\log v_i$ against the median and median absolute
deviation ($\mathrm{MAD}$) of the clean labeled set $\Dl$, $z_i = (\log v_i - \mathrm{med})/\mathrm{MAD}$, and use
$|z_i|$ as the corruption score, so that an image is flagged when its
high-frequency content is atypical in either direction. The score is mapped to a
per-image weight $w_i = \mathrm{sigmoid}\big((z_{\text{cut}} - |z_i|)/\beta\big) \in [0,1]$,
with $z_{\text{cut}} = 3$ and $\beta = 0.5$, where small $w_i$ indicates likely
corruption. Here $z_{\text{cut}}$ is the score at which $w_i = 0.5$, and $\beta$ controls how sharply the weight falls from 1 to 0 around it. It cannot see corruptions that leave high-frequency content unchanged, such as changes of brightness, contrast or saturation; the quality detector below covers those.

\textbf{Two-stage corruption matcher.} Section~\ref{sec:anchor} requires
labeled images corrupted in the same way as the pool. The matcher selects,
from a fixed library of synthetic corruptions, the operator
$(c^\ast, s^\ast)$ whose output most resembles the corrupted part of the
pool. It first identifies that part of the pool, and then narrows the
candidates in two stages: a cheap comparison of image statistics, followed
by a comparison in feature space.

\textbf{Candidates and target set.} Let $\mathcal{C}$ be 18 of the 19 corruption
families of~\citet{michaelis2019imagecorruptions} (glass blur is held out of the library and used only in Section~\ref{sec:beyond}) and
$\mathcal{S} = \{1, \dots, 5\}$ their severities, and let
$\corrupt{c}{s}(\cdot)$ denote the operator for family $c \in \mathcal{C}$
at severity $s \in \mathcal{S}$. For each candidate pair we construct a bank
of corrupted labeled images,
$\mathcal{B}_{c,s} = \{\corrupt{c}{s}(\boldsymbol{x}^l_j)\}_j$.
The target set $\mathcal{T} \subset \Du$ is the part of the pool suspected
to be corrupted. It is chosen by a quality detector that is separate from the sensor, because the sensor reads only high-frequency content and cannot see corruptions such as brightness, contrast or saturation.
Let $\Dl_{\text{cal}} \subset \Dl$ be a calibration sample of clean labeled images. We standardize each of the $d = 6$ quality statistics of Stage One by its median and robust scale ($1.4826\,\mathrm{MAD}$) on $\Dl_{\text{cal}}$, which gives a vector $\boldsymbol{u}(\boldsymbol{x}) \in \mathbb{R}^d$, and score an image by its robust Mahalanobis distance to the clean labeled images,
\begin{equation}
\label{eq:detector}
r(\boldsymbol{x}) = \boldsymbol{u}(\boldsymbol{x})^{\top}\,\Sigma_{\alpha}^{-1}\,\boldsymbol{u}(\boldsymbol{x}),
\qquad
\Sigma_{\alpha} = (1 - \alpha)\,\hat{\Sigma} + \alpha I,
\end{equation}
where $\hat{\Sigma}$ is the covariance of $\boldsymbol{u}$ over $\Dl_{\text{cal}}$ and $\alpha = 0.1$ is a shrinkage weight, so that a few outlying labeled images do not distort it. We work on a random subsample of $N$ pool images. Each subsampled image $\boldsymbol{x}^u_i$ receives an empirical $p$-value against the clean scores, $p_i = \big(1 + |\{\boldsymbol{x} \in \Dl_{\text{cal}} : r(\boldsymbol{x}) \ge r(\boldsymbol{x}^u_i)\}|\big) / \big(|\Dl_{\text{cal}}| + 1\big)$, and we estimate the contamination rate with Storey's method~\citep{storey2002fdr},
\begin{equation}
\label{eq:storey}
\hat\rho = 1 - \frac{|\{i : p_i > \lambda_S\}|}{(1 - \lambda_S)\, N},
\end{equation}
with $\lambda_S = 0.5$, clipped to $[0, 1]$. The $p$-value of a clean image is spread evenly over $[0, 1]$, so about a fraction $1 - \lambda_S$ of the clean images lies above $\lambda_S$, and the shortfall from that count estimates the corrupted fraction. $\mathcal{T}$ is then the $n_{\mathcal{T}} = \mathrm{round}\big(\max(\hat\rho, 0.01)\, N\big)$ subsampled images with the highest $r$, and the reference set $\mathcal{R}$ is the $n_{\mathcal{T}}$ images with the lowest $r$, that is, the pool images that look most like clean labeled images. Comparing $\mathcal{B}_{c,s}$ to $\mathcal{T}$ in
pixel space alone would be dominated by image content rather than
corruption, so we compare them in two stages.

\textbf{Stage One:}
For each image in $\mathcal{B}_{c,s}$ and $\mathcal{T}$ we compute $d = 6$ low-level
quality statistics: noise level~\citep{immerkaer1996noise}, variance of the
Laplacian~\citep{pechpacheco2000diatom}, Tenengrad gradient
energy~\citep{krotkov1988focusing}, contrast, saturation, and brightness. The
two sharpness statistics are taken as $\log(x + 10^{-10})$. We standardize each statistic by its
median and MAD on clean labeled images, and compare the
two groups using the mean per-statistic 1-Wasserstein distance,

\begin{equation}
\label{eq:wasserstein}
  W_1^{\text{avg}}(c,s) = \frac{1}{d}\sum_{j=1}^{d}
  W_1\!\big(\mathcal{B}^{(j)}_{c,s},\, \mathcal{T}^{(j)}\big).
\end{equation}

Here $\mathcal{B}^{(j)}_{c,s}$ and $\mathcal{T}^{(j)}$ are the values of the $j$-th statistic over the two sets, and $W_1$ between two one-dimensional samples is the area between their empirical cumulative distribution functions. The $n_{\text{surv}} = 8$ pairs with the smallest $W_1^{\text{avg}}$ form the survivor set
$\mathcal{G}' \subseteq \mathcal{C} \times \mathcal{S}$.

\textbf{Stage Two:}
For each survivor $(c,s) \in \mathcal{G}'$ we compare the bank and
the target in the feature space of the frozen DINO backbone $f_0(\cdot)$, using Maximum mean discrepancy (MMD)~\citep{gretton2012mmd} with an RBF kernel $\kappa$ on backbone features, written $\kappa(a,b) = \kappa\big(f_0(a), f_0(b)\big)$. MMD measures how far apart two sets of samples are as distributions. It is zero for identical distributions and grows as they differ. 

\begin{equation}
\label{eq:mmd}
\mathrm{MMD}^2(\mathcal{B}_{c,s}, \mathcal{T}) = \mathbb{E}_{a,a'\sim\mathcal{B}_{c,s}}\big[\kappa(a,a')\big] - 2\,\mathbb{E}_{a\sim\mathcal{B}_{c,s},\,b\sim\mathcal{T}}\big[\kappa(a,b)\big] + \mathbb{E}_{b,b'\sim\mathcal{T}}\big[\kappa(b,b')\big].
\end{equation}

Here $\mathbb{E}$ denotes the expectation, which we estimate by averaging over all pairs in the two sets, and $a, a'$ and $b, b'$ are drawn independently from $\mathcal{B}_{c,s}$ and $\mathcal{T}$ respectively. The pair minimizing $\mathrm{MMD}^2$ is the matched corruption
$(c^\ast, s^\ast)$, which is used by all later components. The kernel bandwidth
is set to the median pairwise feature distance between clean labeled images and $\mathcal{R}$ (the median heuristic), so all candidates are compared on one scale. 

We use two stages rather than MMD alone because MMD in backbone space is expensive over the full grid $\mathcal{C} \times \mathcal{S}$, while Stage One is cheap and removes candidates that produce the wrong kind of corruption.

\textbf{Fitted operator.}
\label{sec:fit}
The library contains a fixed set of operators, which can not always properly represent a corruption. If the closest library operator is a poor stand-in, the anchors built from it do not reach the corrupted region of feature space. We measure how far a bank $\mathcal{B}$ is from the target by its excess over a floor,
\begin{equation}
\label{eq:excess}
\delta(\mathcal{B}) = \frac{\mathrm{MMD}^2(\mathcal{B}, \mathcal{T}) - \mathrm{MMD}^2(\Dl, \mathcal{R})}{\mathrm{MMD}^2(\Dl, \mathcal{R})},
\end{equation}
where the floor $\mathrm{MMD}^2(\Dl, \mathcal{R})$ compares clean labeled images with the clean-looking part of the pool, so it is the distance expected between two clean sets, and $\delta$ is measured in units of it (floor units). We fit a new operator only when two independent signals agree: (A) the detector finds a corrupted part, $\hat\rho \ge 0.02$, and (B) the matched operator is far from it, $\delta(\mathcal{B}_{c^\ast,s^\ast}) \ge K$ with $K = 5$. Otherwise the library choice is used unchanged. Requiring both signals prevents a fit on a clean pool, where the target contains no corruption and every library operator is far from it.

The fitted operator $g_\theta$ is built only from appearance operations, so it cannot change an image's class. We use two parameterizations. The parametric operator has 13 parameters (anisotropic Gaussian and motion blur, local pixel displacement as in glass blur~\citep{hendrycks2019benchmarking}, Gaussian and impulse noise, brightness, contrast and saturation) and is fitted with the derivative-free evolution strategy CMA-ES~\citep{hansen2001cmaes}. The free-form operator is differentiable and is fitted by gradient descent with Adam~\citep{kingma2015adam} through the frozen backbone. It applies a free $15 \times 15$ blur kernel, a smooth random warp implemented by differentiable resampling~\citep{jaderberg2015stn}, per-channel noise and the same photometric terms, twice in sequence, either as a single operator ($k = 1$) or as a mixture of two operators ($k = 2$) in which each image is assigned to one component. Every candidate is fitted to the objective of Stage Two,
\begin{equation}
\label{eq:fit}
\theta^\ast = \arg\min_{\theta}\, \mathrm{MMD}^2\big(\{g_\theta(\boldsymbol{x}^l_j)\}_j,\, \mathcal{T}\big),
\end{equation}
which fits a generator to a target sample by minimizing $\mathrm{MMD}^2$, as in generative moment matching networks~\citep{li2015gmmn}, with the corruption operator in the role of the generator. Each image draws the randomness of $g_\theta$ from its own seed, so its corrupted twin is the same in every epoch, as with library operators. Training then uses the candidate with the smallest $\delta$ among the library choice $(c^\ast, s^\ast)$ and the three fits. Section~\ref{sec:beyond} evaluates this rule on pools corrupted by operators outside the library.

\subsection{Extending Anchors into the Corrupted Region}
\label{sec:anchor}

In Section~\ref{sec:estimate}, we estimated the corruption in the unlabeled pool. In this section, we aim to transfer this corruption to give supervision in the corrupted region. We call a corrupted image whose label we trust an anchor, which can supervise a prototype in the corrupted region.

\textbf{Labeled anchor extension.} For each labeled image $\boldsymbol{x}^l_i$ we create a corrupted twin $\corrupt{c^\ast}{s^\ast}(\boldsymbol{x}^l_i)$, a copy corrupted by the matched operator that keeps the label $y^l_i$. The extended labeled set is

\begin{equation}
\label{eq:extend}
  \Dl_{+} = \Dl \cup \big\{\big(\corrupt{c^\ast}{s^\ast}(\boldsymbol{x}^l_i),\, y^l_i\big)\big\}_{i=1}^{N_l},
\end{equation}

and we use $\Dl_{+}$ in place of $\Dl$ throughout Eq.~\eqref{eq:simgcd}, so the twins are ordinary labeled samples in every term. Each known-class prototype is then supervised by both the clean and the corrupted appearance of its class, so that a corrupted unlabeled image of a known class can be assigned to its own class.

\textbf{Novel-class pseudo-anchors.} After a fixed number of epochs $E_{\text{pa}}$, once the classifier has formed initial novel clusters, we generate corrupted twins using the same matched operator to give corrupted images of novel classes supervision as well. We use only the model's most confident guesses, not all novel images.

\emph{Selection:} At epoch $E_{\text{pa}}$ we perform a single inference pass
over $\Du$, producing for each image the class-probability vector $\boldsymbol{p}_i$ of Eq.~\eqref{eq:softmax}. Let
$\mathrm{pred}_i = \arg\max_k p_i^{(k)}$, and let
$m_i$ denote the margin between the largest and second-largest entries of $\boldsymbol{p}_i$. For each novel prototype $k$ we select the $n_{\text{pa}}$
images predicted into that prototype with the largest margins to get the set of pseudo-anchors for novel prototype $k$, $\mathcal{A}_k$,

\begin{equation}
\label{eq:select}
  \begin{gathered}
\mathcal{A}_k = \operatorname{top-}\!n_{\text{pa}}
  \big\{\, m_i \;:\; i \in \Du,\ \mathrm{pred}_i = k \,\big\},
\end{gathered}
\end{equation}

and assign each the pseudo-label $k$. Selection is performed per prototype rather than globally. We set $n_{\text{pa}}$ to one fifth of the average number of
training images per class (1000, 100, 6, 13 and 8 on \cifarten, \cifarhundred, \cub,
\aircraft and \cars).

\emph{Injection.} Each selected image is corrupted with the same operator $(c^\ast, s^\ast)$, and added to the
labeled set carrying its pseudo-label, in the same way as the labeled
twins. Injection works as a step
function, where before $E_{\text{pa}}$ the pseudo-anchors are present with sampling
weight zero, and from $E_{\text{pa}}$ onward they are sampled at a fixed weight
$\alpha_{\text{pa}}$ for the remainder of training, which is the relative chance of drawing a pseudo-anchor to a labeled image. We use $E_{\text{pa}} = 30$
throughout.

\subsection{Trust Curriculum and Paired Alignment}
\label{sec:align}

With anchors in place, we add two components. A trust curriculum uses the
sensor weights of Section~\ref{sec:estimate} to down-weight likely corrupted
images while prototypes form, then gradually restores them once anchors can
make use of them. A paired alignment loss pulls each corrupted twin back
toward its own clean source when the two drift apart, which cross-entropy alone
does not require.

\textbf{Trust curriculum.} The sensor scores are used to weight the unlabeled
pool so that likely corrupted images contribute less while prototypes are
forming. Two considerations conflict here. Early in training, we expect corrupted images to be harmful, since prototypes have not formed and a corrupted image assigned to a novel prototype could start the drift described in Section~\ref{sec:motivation}. Later in training, we expect the same corrupted images to be useful, since anchors then exist in the corrupted region and a corrupted image is a valid training example of its class, so suppressing it discards data the anchors were built to make usable.

We resolve this by using a ramping schedule for weighting. Let $w_i \in [0,1]$ be the sensor weight of unlabeled image $i$, let $t \in [0,1]$ be the fraction of training completed, and let $\gamma(t) = \min(1, t/\eta)$, which ramps linearly from $0$ to $1$ over the first $\eta$ fraction of training. The effective weight is

\begin{equation}
\label{eq:trust}
  \omega_i(t) = \big(1 - \gamma(t)\big)\, w_i + \gamma(t),
\end{equation}

so the pool is weighted by the raw sensor scores at $t = 0$ and uniformly from
$t = \eta$ onward. The weights $\omega_i(t)$ only multiply the per-image self-distillation loss of unlabeled images, while the mean-entropy term, labeled images, twins and pseudo-anchors are unweighted.

We use $\eta = 0.5$ by default, meaning the sensor's influence decays to zero
halfway through training and the model is trained on the full pool at uniform
weight for the second half. The sensor is used to protect
prototypes while they are forming, and once the anchors of
Section~\ref{sec:anchor} are in place, permanently down-weighting part of the pool would reduce the training data without a corresponding benefit.

\textbf{Paired alignment loss.} For a labeled image $\boldsymbol{x}_i$ with
corrupted twin
$\tilde{\boldsymbol{x}}_i = \corrupt{c^\ast}{s^\ast}(\boldsymbol{x}_i)$, we
penalize the cosine similarity between the twin's feature and its own clean
source only when it falls below a threshold $\delta$, with the gradient stopped
on the clean side:
\begin{equation}
\label{eq:align}
  \mathcal{L}_{\text{align}} = \frac{1}{n}\sum_{i=1}^{n}
  \max\!\Big(0,\ \delta - \cos\!\big(f(\tilde{\boldsymbol{x}}_i),\ \sg\big(f(\boldsymbol{x}_i)\big)\big) \Big).
\end{equation}
Here $n$ is the number of labeled pairs in the batch, and $\sg(\cdot)$ is the stop-gradient operator, which passes its input forward unchanged but blocks gradients from flowing through it. Both images of a pair receive the same augmentation draw, so corruption is the
only difference between them. Pairs are drawn from labeled originals only;
pseudo-anchors are not paired, since a pool image has no clean source. We fix
$\delta = 0.5$ in advance. A pair that already agrees receives no gradient, so
the term is a safeguard against drift rather than a constant pull. The
stop-gradient makes the clean feature a fixed target. Without it, the loss could also be reduced by moving the clean feature toward the corrupted one, which would damage the clean features the method relies on. The full objective is

\begin{equation}
\label{eq:total}
  \mathcal{L} = \mathcal{L}_{\text{SimGCD}} + \lambda_{\text{align}}\,\mathcal{L}_{\text{align}}.
\end{equation}

\section{Experiments}
\label{sec:exp}

\subsection{Experimental Setup}
\label{sec:setup}

\textbf{Datasets and splits.} We evaluate on two generic datasets, \cifarten and \cifarhundred~\citep{krizhevsky2009learning}, and three fine-grained datasets, \cub~\citep{WahCUB_200_2011}, \cars~\citep{Krause_2013_ICCV_Workshops} and \aircraft~\citep{maji13fine-grained}, for which we use the known and novel class splits of the Semantic Shift Benchmark~\citep{vaze2022ssb}. For each dataset we sub-sample $|\Yk|$ classes as known classes and treat the rest as novel. We then sub-sample 50\% of the images of each known class to form $\Dl$, and concatenate the remainder with all images of the novel classes to form $\Du$. The default $|\Yk|$ for \cifarten, \cifarhundred, \cub, \cars and \aircraft is 5, 80, 100, 98 and 50.

\textbf{Contamination.} The corruption operators are those of the common-corruption benchmark~\citep{hendrycks2019benchmarking}, in our replicated version of the imagecorruptions package~\citep{michaelis2019imagecorruptions}. Each run uses one corruption family. A fraction $\rho$ of $\Du$ is selected uniformly at random, and each selected image is replaced by its corrupted version at severity~5, the strongest level the benchmark defines. The labeled set $\Dl$ is never corrupted. We apply the operators ourselves to the training split of each dataset, which lets us corrupt a controlled fraction of the pool and extend the protocol to the three fine-grained datasets, for which no corrupted benchmark is distributed.

\looseness-1\textbf{Evaluation.} Following SimGCD~\citep{wen2023simgcd}, we match predicted clusters to ground-truth classes with the Hungarian algorithm over all $|\Yk| + |\Yn|$ classes, and report accuracy on all of $\Du$ (All), on known classes (Known) and on novel classes (Novel). We also split $\Du$ into its clean and corrupted images and report Known and Novel on each part, writing $\text{Known}_c$ and $\text{Novel}_c$ for the corrupted part. These splits use corruption membership, which the method never sees, and serve only as a diagnostic. Every result is a single run.

\textbf{Implementation details.} Following common practice for GCD, SimGCD and \method use a ViT-B/16 backbone initialized with DINO~\citep{caron2021dino}. Following SimGCD~\citep{wen2023simgcd}, we fine-tune only the last block and train for 200 epochs with batch size 128 and an initial learning rate of 0.1. Pseudo-anchors are injected at $E_{\text{pa}} = 30$ with $n_{\text{pa}}$ set to one fifth of the training images per class (Section~\ref{sec:anchor}) and sampling weight $\alpha_{\text{pa}} = 1$. We set $\lambda_{\text{align}} = 1$ and $\delta = 0.5$.

\subsection{Main Results}
\label{sec:exp-main}

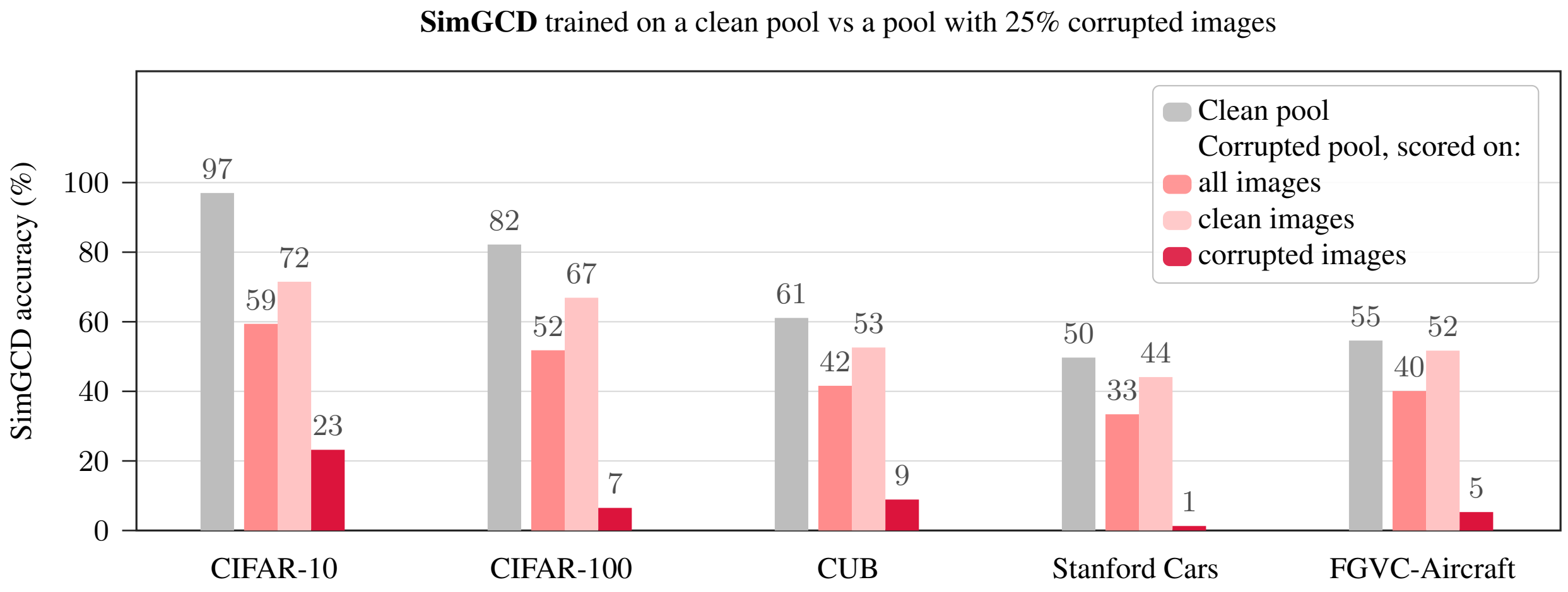
\begin{figure}[!htb]
  \centering
  \begin{tikzpicture}
  \begin{axis}[paper, ybar, bar width=10pt, width=\textwidth, height=6.4cm,
    ymin=0, ymax=132, ytick={0,20,...,100}, ylabel={SimGCD accuracy (\%)}, xtick={0,...,4},
    xticklabels={CIFAR-10, CIFAR-100, CUB, Stanford Cars, FGVC-Aircraft},
    xtick style={draw=none}, xmajorgrids=false, enlarge x limits=0.12,
    title={\textbf{SimGCD} trained on a clean pool vs a pool with 25\% corrupted images},
    nodes near coords, nodes near coords style={font=\footnotesize, color=black!70},
    point meta=rawy, every node near coord/.append style={/pgf/number format/.cd, fixed, precision=0},
    legend style={at={(0.985,0.97)}, anchor=north east, legend columns=1},
    legend image code/.code={\draw[#1, draw=none] (0cm,-0.1cm) rectangle (0.3cm,0.1cm);}]
    \addplot[fill=cBase, draw=none, bar shift=-17pt] coordinates {(0,97.0) (1,82.2) (2,61.1) (3,49.7) (4,54.6)};
    \addlegendimage{empty legend}
    \addplot[fill=cSalmon, draw=none, bar shift=-4pt] coordinates {(0,59.4) (1,51.8) (2,41.6) (3,33.4) (4,40.1)};
    \addplot[fill=cSalmonLight, draw=none, bar shift=6pt] coordinates {(0,71.5) (1,66.9) (2,52.6) (3,44.1) (4,51.7)};
    \addplot[fill=cSim, draw=none, bar shift=16pt] coordinates {(0,23.2) (1,6.5) (2,8.9) (3,1.3) (4,5.3)};
    \legend{Clean pool, {Corrupted pool, scored on:}, all images, clean images, corrupted images}
  \end{axis}
  \end{tikzpicture}
  \caption{SimGCD trained on a clean pool ($\rho = 0$, gray) vs a pool with $\rho = 0.25$ (red), protocol of Section~\ref{sec:setup}, severity~5, averaged over the families evaluated on each dataset (Appendix~\ref{app:perfamily}). The three red bars are one training run, scored on all of $\Du$, on its clean images and on its corrupted images. The last two splits use corruption membership, which the method never sees. Clean-pool numbers in Table~\ref{tab:vanilla}, contaminated-pool numbers in Table~\ref{tab:main}.}
  \label{fig:contamination}
\end{figure}

Contamination costs SimGCD a large share of its accuracy, as expected in Section~\ref{sec:motivation} (Figure~\ref{fig:contamination}). With a quarter of the pool corrupted, All accuracy falls by 14.5 to 37.6 points across the five datasets, and the corrupted images are almost never classified correctly (1\% to 23\%). The loss also reaches the clean images. If every corrupted image were misclassified and every clean image kept its clean-pool accuracy, accuracy would equal $(1-\rho)$ times its clean-pool value. Novel accuracy falls below this floor on all five datasets, by 20.6 points on \cifarten, 22.8 on \cifarhundred, 6.9 on \cub, 3.6 on \cars and 0.7 on \aircraft (Tables~\ref{tab:main} and~\ref{tab:vanilla}). Corruption therefore damages the novel prototypes themselves.

\method improves All accuracy over SimGCD and HiLo on every dataset (Table~\ref{tab:main}). The gain over SimGCD is 32.9 points on \cifarten, 19.8 on \cifarhundred, 13.6 on \cub, 15.7 on \cars and 6.3 on \aircraft, which recovers 87\%, 65\%, 70\%, 96\% and 43\% of the All accuracy that SimGCD loses to contamination. The gain comes from the corrupted images. SimGCD and HiLo classify almost none of them correctly ($\text{Known}_c$ at most 20.9), while \method reaches a $\text{Known}_c$ of 50.4 to 83.2 and a $\text{Novel}_c$ of 31.6 to 75.1. On clean images, \method stays within 0.4 points of SimGCD or improves on it, except for clean Novel accuracy on \aircraft, which falls by 5.0 points.

\begin{table}[!htb]
\centering
\small
\caption{Accuracy (\%) on the unlabeled pool $\Du$ with $\rho = 0.25$ of the pool corrupted at severity~5, averaged over the corruption families evaluated on each dataset (Appendix~\ref{app:perfamily}). HiLo is the domain-shift GCD baseline. Clean and Corrupted columns stratify $\Du$ by whether an image was corrupted. Whole-pool columns are the standard GCD metrics. Single run per cell.}

\label{tab:main}
\begin{tabular*}{\textwidth}{@{\extracolsep{\fill}}ll cc cc ccc}
\toprule
& & \multicolumn{2}{c}{\textbf{Clean subset}} & \multicolumn{2}{c}{\textbf{Corrupted subset}} & \multicolumn{3}{c}{\textbf{Whole pool}} \\
\cmidrule(lr){3-4} \cmidrule(lr){5-6} \cmidrule(lr){7-9}
\textbf{Dataset} & \textbf{Method} & \textbf{Known} & \textbf{Novel} & \textbf{Known} & \textbf{Novel} & \textbf{All} & \textbf{Known} & \textbf{Novel} \\
\midrule
CIFAR-10 & SimGCD & \textbf{96.3} & 59.2 & 0.0 & 35.1 & 59.4 & 71.8 & 53.2 \\
 & HiLo & 93.1 & \textbf{98.8} & 20.9 & 0.0 & 74.4 & 74.7 & 74.3 \\
 & \method & 95.9 & 97.8 & \textbf{83.2} & \textbf{75.1} & \textbf{92.3} & \textbf{92.7} & \textbf{92.2} \\
\midrule
CIFAR-100 & SimGCD & 76.1 & 48.5 & 3.9 & 11.6 & 51.8 & 57.9 & 39.4 \\
& HiLo & 76.4 & 59.0 & 1.9 & 9.8 & 54.1 & 57.6 & 46.9 \\
 & \method & \textbf{81.9} & \textbf{67.6} & \textbf{61.0} & \textbf{43.3} & \textbf{71.6} & \textbf{76.7} & \textbf{61.6} \\
\midrule
CUB & SimGCD & 65.2 & 46.3 & 3.1 & 11.8 & 41.6 & 49.5 & 37.7 \\
& HiLo & 64.3 & 46.9 & 4.0 & 11.1 & 41.7 & 49.1 & 38.0 \\
 & \method & \textbf{65.4} & \textbf{55.9} & \textbf{53.6} & \textbf{38.5} & \textbf{55.2} & \textbf{62.4} & \textbf{51.5} \\
\midrule
Stanford Cars & SimGCD & 62.9 & 35.3 & 0.2 & 1.9 & 33.4 & 46.4 & 27.2 \\
& HiLo & 67.2 & 39.8 & 0.6 & 4.0 & 37.1 & 49.7 & 31.1 \\
 & \method & \textbf{72.3} & \textbf{41.7} & \textbf{61.8} & \textbf{31.6} & \textbf{49.1} & \textbf{69.6} & \textbf{39.2} \\
\midrule
FGVC-Aircraft & SimGCD & 59.7 & \textbf{47.6} & 1.0 & 7.5 & 40.1 & 45.1 & 37.6 \\
& HiLo & 48.6 & 47.3 & 1.8 & 6.7 & 37.1 & 37.0 & 37.1 \\
 & \method & \textbf{60.7} & 42.6 & \textbf{50.4} & \textbf{33.9} & \textbf{46.4} & \textbf{58.1} & \textbf{40.5} \\
\bottomrule
\end{tabular*}
\end{table}

The gains are distributed differently on generic and fine-grained datasets. On \cifarten and \cifarhundred, Novel accuracy improves more than Known accuracy, by 39.0 and 22.2 points against 20.9 and 18.8. On \cub, \cars and \aircraft, Known accuracy improves by 12.9, 23.2 and 13.0 points and Novel accuracy by 13.8, 12.0 and 2.9.

The families evaluated differ between datasets, so the means in Table~\ref{tab:main} are not taken over the same families. Per family (Appendix~\ref{app:perfamily}), the All gain is positive in 38 of the 39 combinations of dataset and family, and in the remaining one, shot noise on \aircraft, All accuracy is unchanged. Gaussian noise is the only family evaluated on all five datasets, and its gains are 32.9, 19.0, 10.6, 11.0 and 3.1 points. The smallest gains are on the noise families of \aircraft, where \method recovers the corrupted images but loses part of the clean Novel accuracy.

\method keeps the accuracy of SimGCD when the pool is clean. With $\rho = 0$ it stays within 2.0 points of the All accuracy of SimGCD on every dataset and is above it on four of the five (Table~\ref{tab:vanilla}).

\begin{figure}[!htb]
  \centering
  \begin{tikzpicture}
  \begin{axis}[paper, name=p0, width=0.52\textwidth, height=5cm, title={CIFAR-100},
    xtick={0,...,5}, xticklabels={0,0.05,0.1,0.25,0.5,1}, xlabel={Contamination rate $\rho$},
    xmin=-0.3, xmax=5.3, ymin=0, ymax=100, ylabel={All accuracy (\%)},
    legend style={at={(0.03,0.04)}, anchor=south west, legend columns=1}]
    \addplot[color=cSim, mark=square*] coordinates {(0,82.2) (1,72.8) (2,68.6) (3,50.7) (4,34.2) (5,12.4)};
    \addplot[color=cHiLo, mark=triangle*, mark options={draw=cHiLoEdge, line width=0.5pt}] coordinates {(0,77.6) (1,73.6) (2,67.3) (3,56.9) (4,39.4) (5,3.6)};
    \addplot[color=cOurs, mark=*] coordinates {(0,80.2) (1,75.2) (2,75.0) (3,69.7) (4,64.6) (5,44.0)};
    \legend{SimGCD, HiLo, \method}
  \end{axis}
  \begin{axis}[paper, name=p1, at={(p0.east)}, xshift=0.9cm, anchor=west, width=0.52\textwidth, height=5cm, title={CUB},
    xtick={0,...,5}, xticklabels={0,0.05,0.1,0.25,0.5,1}, xlabel={Contamination rate $\rho$},
    xmin=-0.3, xmax=5.3, ymin=0, ymax=70]
    \addplot[color=cSim, mark=square*] coordinates {(0,61.1) (1,56.3) (2,52.2) (3,40.7) (4,25.2) (5,13.7)};
    \addplot[color=cHiLo, mark=triangle*, mark options={draw=cHiLoEdge, line width=0.5pt}] coordinates {(0,59.0) (1,54.7) (2,51.4) (3,41.9) (4,26.2) (5,12.8)};
    \addplot[color=cOurs, mark=*] coordinates {(0,63.0) (1,60.0) (2,57.1) (3,51.3) (4,46.1) (5,30.0)};
  \end{axis}
  \end{tikzpicture}
  \caption{All accuracy (\%) as the contamination rate $\rho$ grows, Gaussian noise at severity~5. The $\rho$ axis is not to scale. The $\rho = 0$ points are the clean-pool runs of Table~\ref{tab:vanilla}. At $\rho = 1$, All accuracy is 12.4, 3.6 and 44.0 on \cifarhundred and 13.7, 12.8 and 30.0 on \cub for SimGCD, HiLo and \method.}
  \label{fig:rho}
\end{figure}
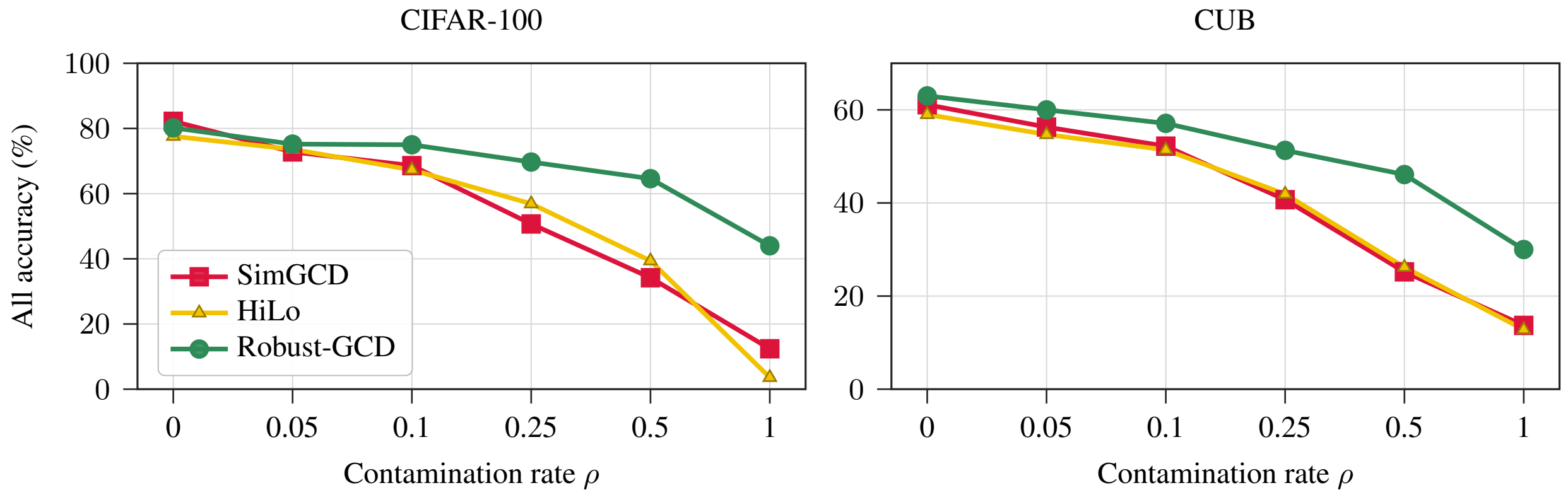

The advantage of \method grows with the contamination rate (Figure~\ref{fig:rho}). On \cifarhundred with Gaussian noise, the All gain over SimGCD rises from 2.4 points at $\rho = 0.05$ to 31.6 at $\rho = 1$, and on \cub from 3.7 to 16.3. At $\rho = 1$, where no clean unlabeled image remains, SimGCD and HiLo fall to between 3.6 and 13.7 All accuracy, while \method keeps 44.0 and 30.0.

\subsection{Ablation and Analysis}
\label{sec:exp-analysis}

We remove or replace each component of \method in turn on \cifarhundred with Gaussian noise, severity~5 and $\rho = 0.25$ (Tables~\ref{tab:abl-operator}, \ref{tab:ablation} and~\ref{tab:abl-onoff}), then test corruptions outside the matcher's library and examine where corrupted images go in feature space. Besides All, Known and Novel we report $\text{Known}_c$ and $\text{Novel}_c$, because the corrupted images are the ones the method exists to fix, so a removed component should show there first.

\textbf{Corruption matcher.} Anchors help only when they resemble the corrupted images. The matcher recovers the true operator on this pool, so the matched row equals the oracle row (Table~\ref{tab:abl-operator}), which uses the true family and severity and serves as a reference. Applying the right family at the wrong severity ($s = 1$), or the wrong family (Gaussian blur on a noisy pool), drops $\text{Known}_c$ to 2.3 and 0.0 and costs 10.8 and 11.9 points of All accuracy. Appendix~\ref{app:estimation} reports $\hat\rho$ and the operator chosen by each stage for every pool.

\textbf{Estimated against random operators.} Estimating the operator matters most for known classes (Table~\ref{tab:abl-operator}). We replace the estimated operator by a random one for twins, pseudo-anchors and alignment pairs, on Gaussian noise and on fog. With a random family, each image receives one of the 18 library families at severity~5, fixed for the whole run. With a random family and severity, the severity is also drawn per image, from 1 to 5. A random family costs 10.5 points of $\text{Known}_c$ on Gaussian noise and 16.4 on fog, and a random severity as well costs 24.0 and 32.0. $\text{Novel}_c$ barely changes with a random family ($+2.5$ and $-1.2$) and falls by 4.3 and 7.2 when the severity is also random, so for novel classes the severity matters more than the family. No run collapses (95 to 99 of 100 clusters).

\begin{table}[!htb]
\centering
\small
\caption{Anchors built with other corruption operators, \cifarhundred, severity~5, $\rho = 0.25$. Each row replaces only the operator used for twins, pseudo-anchors and alignment pairs. Oracle uses the true family and severity and serves as a reference. Random family: one of the 18 library families per image at severity~5. Random family and severity: the severity is also drawn per image, from 1 to 5. Last row of each pool: the operator estimated by the matcher, as used by \method. All, Known, Novel: whole pool. $\text{Known}_c$, $\text{Novel}_c$: corrupted images only. Bold: best in each pool. Single run per row.}
\label{tab:abl-operator}
\begin{tabular*}{\textwidth}{@{\extracolsep{\fill}}l ccc cc}
\toprule
\textbf{Operator} & \textbf{All} & \textbf{Known} & \textbf{Novel} & \textbf{$\text{Known}_c$} & \textbf{$\text{Novel}_c$} \\
\midrule
\multicolumn{6}{l}{\textit{Gaussian noise pool}} \\
Wrong family (blur) & 57.8 & 62.0 & 49.4 & 0.0 & 7.1 \\
Wrong severity ($s = 1$) & 58.9 & 61.3 & 54.0 & 2.3 & 5.7 \\
Random family and severity & 65.3 & 70.6 & 54.8 & 34.5 & 32.5 \\
Random family & 68.9 & 73.6 & \textbf{59.6} & 48.0 & \textbf{39.3} \\
Oracle (true $c$, $s$) & \textbf{69.7} & \textbf{75.3} & 58.7 & \textbf{58.5} & 36.8 \\
Matched $(c^\ast, s^\ast)$, \method & \textbf{69.7} & \textbf{75.3} & 58.7 & \textbf{58.5} & 36.8 \\
\midrule
\multicolumn{6}{l}{\textit{Fog pool}} \\
Random family and severity & 61.9 & 66.2 & 53.3 & 17.6 & 25.7 \\
Random family & 65.0 & 70.0 & 55.0 & 33.2 & 31.7 \\
Matched $(c^\ast, s^\ast)$, \method & \textbf{68.3} & \textbf{72.7} & \textbf{59.4} & \textbf{49.6} & \textbf{32.9} \\
\bottomrule
\end{tabular*}
\end{table}

\textbf{Anchor extension.} Labeled anchors fix the known classes, and pseudo-anchors are needed for the novel classes. Labeled twins raise $\text{Known}_c$ from 2.6 to 61.0 (Table~\ref{tab:ablation}). They also lower $\text{Novel}_c$ from 8.2 to 0.0, which suggests that corrupted novel images are drawn into the known classes once those classes have anchors in the corrupted region. Pseudo-anchors restore $\text{Novel}_c$ to 36.8. Twins without labels gain far less than labeled twins (All 53.5 against 67.7), so the label carried by a twin, not the extra corrupted images, is what helps.

\begin{table}[!htb]
\centering
\small
\caption{Anchor ablation of \method on \cifarhundred, Gaussian noise, severity~5, $\rho = 0.25$. Rows add one kind of anchor at a time, from SimGCD, which has none, to \method in the last row. Twins: corrupted copies of labeled images, made with the matched operator. Columns as in Table~\ref{tab:abl-operator}. Bold: best in each column. Single run per row.}
\label{tab:ablation}
\begin{tabular*}{\textwidth}{@{\extracolsep{\fill}}l ccc cc}
\toprule
\textbf{Anchors} & \textbf{All} & \textbf{Known} & \textbf{Novel} & \textbf{$\text{Known}_c$} & \textbf{$\text{Novel}_c$} \\
\midrule
None (SimGCD) & 50.7 & 56.7 & 38.6 & 2.6 & 8.2 \\
Twins without labels & 53.5 & 57.8 & 44.9 & 28.7 & 12.4 \\
Labeled twins & 67.7 & \textbf{75.4} & 52.3 & \textbf{61.0} & 0.0 \\
Labeled twins $+$ pseudo-anchors (\method) & \textbf{69.7} & 75.3 & \textbf{58.7} & 58.5 & \textbf{36.8} \\
\bottomrule
\end{tabular*}
\end{table}

\textbf{Paired alignment.} Alignment has no consistent effect on accuracy, and we keep it as a safeguard. Cross-entropy asks that an image and its corrupted twin both be classified correctly, and the alignment loss also asks that their features be close (Section~\ref{sec:align}). We turn it off on Gaussian noise and on three blur families (Table~\ref{tab:abl-onoff}). Removing it lowers $\text{Novel}_c$ by 2.8 points on Gaussian noise, 4.2 on zoom blur and 6.9 on defocus blur, and raises it by 3.2 on motion blur. No run collapses. On defocus and motion blur, corrupted images spread over 97 to 99 of the 100 clusters with or without the loss, and by the end of training the hinge is inactive on almost every pair.

\textbf{Trust curriculum.} We test the curriculum at $\rho = 0.5$, where half of the unlabeled images that shape the novel prototypes early in training are corrupted, so its protection should matter most. The ramp reaches 64.6 All accuracy and 56.0 Novel accuracy, against 61.4 and 44.9 for uniform weighting (Table~\ref{tab:abl-onoff}). Known accuracy is about the same (68.9 against 69.6), and the gain is on novel classes, including corrupted novel images ($\text{Novel}_c$ 41.6 against 32.2). This is consistent with the purpose of the curriculum, since the known prototypes are held in place by labeled images and the novel prototypes are not.

\begin{table}[!htb]
\centering
\small
\caption{Switching off one safeguard of \method on \cifarhundred, severity~5. Each pair differs only in the named component. Alignment pairs use $\rho = 0.25$, and the trust pair uses Gaussian noise at $\rho = 0.5$. The second row of each pair is \method. Columns as in Table~\ref{tab:abl-operator}. Bold: better of each pair. Single run per row.}
\label{tab:abl-onoff}
\begin{tabular*}{\textwidth}{@{\extracolsep{\fill}}l ccc cc}
\toprule
\textbf{Setting} & \textbf{All} & \textbf{Known} & \textbf{Novel} & \textbf{$\text{Known}_c$} & \textbf{$\text{Novel}_c$} \\
\midrule
\multicolumn{6}{l}{\textit{Alignment loss}} \\
Gaussian noise, off & \textbf{69.9} & \textbf{76.2} & 57.4 & \textbf{58.7} & 34.0 \\
Gaussian noise, on & 69.7 & 75.3 & \textbf{58.7} & 58.5 & \textbf{36.8} \\
\addlinespace[3pt]
Zoom blur, off & 77.6 & 80.2 & 72.3 & 72.8 & 66.2 \\
Zoom blur, on & \textbf{79.7} & \textbf{80.5} & \textbf{78.1} & \textbf{73.2} & \textbf{70.4} \\
\addlinespace[3pt]
Defocus blur, off & 77.0 & \textbf{80.4} & 70.3 & 70.7 & 63.3 \\
Defocus blur, on & \textbf{78.5} & 79.9 & \textbf{75.6} & \textbf{71.2} & \textbf{70.2} \\
\addlinespace[3pt]
Motion blur, off & \textbf{77.4} & 79.6 & \textbf{73.1} & 70.7 & \textbf{65.2} \\
Motion blur, on & 77.1 & \textbf{79.8} & 71.7 & \textbf{70.9} & 62.0 \\
\midrule
\multicolumn{6}{l}{\textit{Trust curriculum, $\rho = 0.5$}} \\
None (uniform weights) & 61.4 & \textbf{69.6} & 44.9 & \textbf{58.8} & 32.2 \\
Ramp, $\eta = 0.5$ & \textbf{64.6} & 68.9 & \textbf{56.0} & 57.5 & \textbf{41.6} \\
\bottomrule
\end{tabular*}
\end{table}

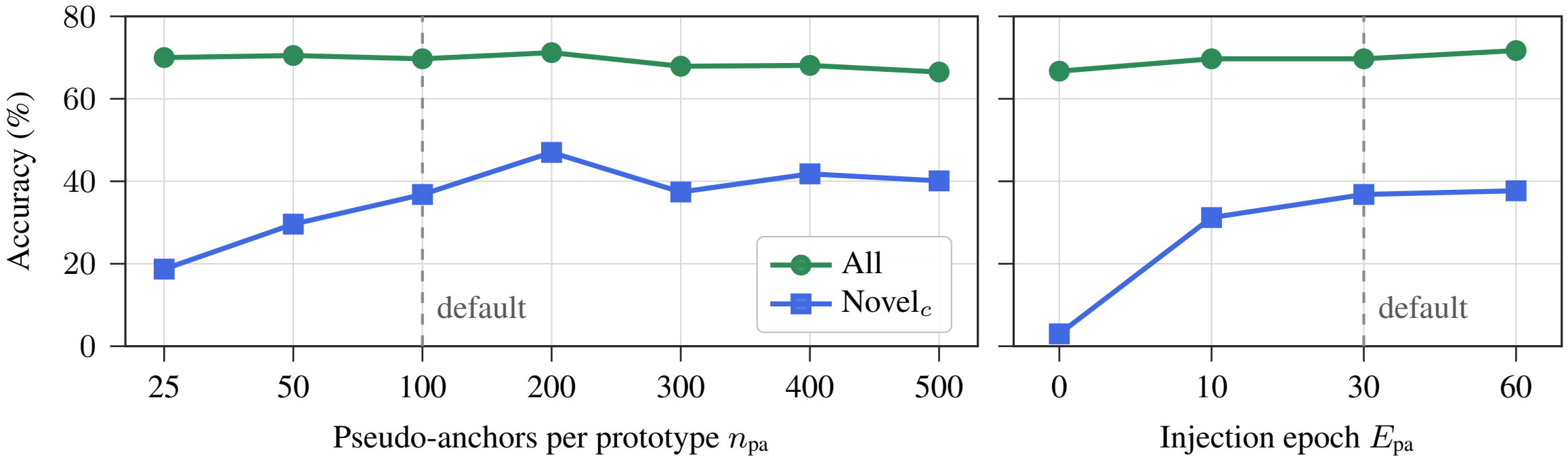
\begin{figure}[!htb]
  \centering
  \begin{tikzpicture}
  \begin{axis}[paper, name=q0, width=0.6\textwidth, height=4.8cm, xtick={0,...,6}, xticklabels={25,50,100,200,300,400,500},
    xlabel={Pseudo-anchors per prototype $n_{\text{pa}}$}, xmin=-0.3, xmax=6.3, ymin=0, ymax=80, ylabel={Accuracy (\%)},
    legend style={at={(0.97,0.04)}, anchor=south east, legend columns=1}]
    \draw[cRef, dashed, line width=0.8pt] (axis cs:2,0) -- (axis cs:2,80) node[pos=0.04, anchor=south west, font=\footnotesize, black!65] {default};
    \addplot[color=cOurs, mark=*] coordinates {(0,70.0) (1,70.5) (2,69.7) (3,71.2) (4,67.9) (5,68.1) (6,66.5)};
    \addplot[color=cAlt, mark=square*] coordinates {(0,18.7) (1,29.6) (2,36.8) (3,47.0) (4,37.4) (5,41.8) (6,40.1)};
    \legend{All, $\text{Novel}_c$}
  \end{axis}
  \begin{axis}[paper, name=q1, at={(q0.east)}, xshift=0.35cm, anchor=west, yticklabels={}, width=0.42\textwidth, height=4.8cm,
    xtick={0,...,3}, xticklabels={0,10,30,60}, xlabel={Injection epoch $E_{\text{pa}}$}, xmin=-0.3, xmax=3.3, ymin=0, ymax=80]
    \draw[cRef, dashed, line width=0.8pt] (axis cs:2,0) -- (axis cs:2,80) node[pos=0.04, anchor=south west, font=\footnotesize, black!65] {default};
    \addplot[color=cOurs, mark=*] coordinates {(0,66.7) (1,69.7) (2,69.7) (3,71.7)};
    \addplot[color=cAlt, mark=square*] coordinates {(0,3.0) (1,31.2) (2,36.8) (3,37.7)};
  \end{axis}
  \end{tikzpicture}
  \caption{Pseudo-anchor settings on \cifarhundred, Gaussian noise, severity~5, $\rho = 0.25$, one run per point. All: whole pool. $\text{Novel}_c$: corrupted images of novel classes. Horizontal axes are not to scale.}
  \label{fig:hparams}
\end{figure}

\textbf{Pseudo-anchor settings.} All accuracy varies little with the pseudo-anchor settings, while $\text{Novel}_c$ depends on both (Figure~\ref{fig:hparams}). $n_{\text{pa}}$ is the number of pseudo-anchors each novel class receives. Too few give little supervision, while too many reach into low-margin guesses, which are wrong more often and are treated as ground truth by the supervised loss. $\text{Novel}_c$ rises from 18.7 at $n_{\text{pa}} = 25$ to 47.0 at $n_{\text{pa}} = 200$ and falls beyond it, while All accuracy stays between 66.5 and 71.2. $E_{\text{pa}}$ is the epoch at which pseudo-anchors are injected. Injecting at epoch~0, when the classifier's guesses are close to random, leaves $\text{Novel}_c$ at 3.0, while from epoch~10 onward it lies between 31.2 and 37.7.

\textbf{Corruptions outside the library.}
\label{sec:beyond}
\method needs an operator close to the corruption, not the exact one. Every pool above was corrupted by an operator in the matcher's library. We test three that are not, on \cifarhundred and \cub with $\rho = 0.25$ (Table~\ref{tab:beyond}): severity drawn per image from $\{3, 4, 5\}$, glass blur, which is held out of the library, and a pool whose corrupted images are half Gaussian noise and half defocus blur. Mixed severity behaves like an in-library pool (All $+19.6$ and $+14.2$ over SimGCD). The other two pools trigger the fitted operator of Section~\ref{sec:fit}, and Table~\ref{tab:fitted-excess} gives the excess of each candidate.

\begin{table}[!htb]
\centering
\small
\caption{Pools outside the library (accuracy, \%, $\rho = 0.25$, one run per row). Columns as in Table~\ref{tab:pf-cifar100}. Library: the matcher's library choice. Fitted: the operator chosen in Table~\ref{tab:fitted-excess}.}
\label{tab:beyond}
\begin{tabular*}{\textwidth}{@{\extracolsep{\fill}}lll ccc cc cc}
\toprule
 & & & \multicolumn{3}{c}{\textbf{All images}} & \multicolumn{2}{c}{\textbf{Clean}} & \multicolumn{2}{c}{\textbf{Corrupted}} \\
\cmidrule(lr){4-6}\cmidrule(lr){7-8}\cmidrule(lr){9-10}
\textbf{Dataset} & \textbf{Pool} & \textbf{Method} & \textbf{All} & \textbf{Known} & \textbf{Novel} & \textbf{Known} & \textbf{Novel} & \textbf{Known} & \textbf{Novel} \\
\midrule
CIFAR-100 & Mixed severity & SimGCD & 50.3 & 57.3 & 36.3 & 76.2 & 45.4 & 1.4 & 8.1 \\
 & & \method & \textbf{69.9} & \textbf{76.9} & \textbf{56.0} & \textbf{82.2} & \textbf{61.4} & \textbf{61.0} & \textbf{39.4} \\
\addlinespace[2pt]
 & Glass blur & SimGCD & 51.3 & \textbf{58.9} & 36.1 & \textbf{77.8} & 41.1 & 2.9 & \textbf{20.7} \\
 & & \method, library & \textbf{53.7} & 58.1 & \textbf{44.8} & 76.9 & \textbf{53.9} & 2.6 & 16.8 \\
 & & \method, fitted & 53.4 & 58.8 & 42.6 & 77.1 & 51.3 & \textbf{4.6} & 15.6 \\
\addlinespace[2pt]
 & Noise + defocus & SimGCD & 52.6 & 58.8 & 40.1 & 77.3 & 49.2 & 4.0 & 12.1 \\
 & & \method, library & 61.6 & 68.1 & 48.7 & 81.6 & 54.3 & 28.1 & 31.7 \\
 & & \method, fitted & \textbf{68.9} & \textbf{72.5} & \textbf{61.8} & \textbf{82.5} & \textbf{68.0} & \textbf{42.8} & \textbf{42.5} \\
\midrule
CUB & Mixed severity & SimGCD & 41.7 & 51.8 & 36.7 & \textbf{68.4} & 44.8 & 2.4 & 12.3 \\
 & & \method & \textbf{55.9} & \textbf{63.6} & \textbf{51.9} & 66.5 & \textbf{56.6} & \textbf{55.2} & \textbf{38.0} \\
\addlinespace[2pt]
 & Glass blur & SimGCD & 41.0 & 50.5 & 36.3 & \textbf{66.6} & 45.2 & 2.6 & 9.5 \\
 & & \method, library & 41.8 & 48.7 & 38.4 & 65.1 & 49.5 & 0.0 & 5.0 \\
 & & \method, fitted & \textbf{49.8} & \textbf{57.8} & \textbf{45.9} & 61.2 & \textbf{53.4} & \textbf{47.8} & \textbf{23.2} \\
\addlinespace[2pt]
 & Noise + defocus & SimGCD & 40.8 & 51.0 & 35.7 & \textbf{67.7} & 44.2 & 1.3 & 10.0 \\
 & & \method, library & 46.9 & 56.5 & 42.0 & 66.0 & 50.3 & 28.4 & 17.3 \\
 & & \method, fitted & \textbf{52.3} & \textbf{60.2} & \textbf{48.3} & 64.0 & \textbf{55.0} & \textbf{48.8} & \textbf{28.1} \\
\bottomrule
\end{tabular*}
\end{table}

\begin{table}[!htb]
\centering
\small
\caption{Excess of each candidate over the floor $\mathrm{MMD}^2(\Dl, \mathcal{R})$, in floor units (lower is closer). Bold: the candidate used for training, the one with the smallest excess (Section~\ref{sec:fit}). Noise + defocus: half Gaussian noise, half defocus blur, both severity~5.}
\label{tab:fitted-excess}
\begin{tabular*}{\textwidth}{@{\extracolsep{\fill}}ll cccc}
\toprule
\textbf{Dataset} & \textbf{Pool} & \textbf{Library} & \textbf{Parametric} & \textbf{Free-form, $k = 1$} & \textbf{Free-form, $k = 2$} \\
\midrule
CIFAR-100 & Glass blur & 35.72 & 15.22 & 22.16 & \textbf{15.02} \\
 & Noise + defocus & 27.22 & 21.80 & 22.94 & \textbf{0.15} \\
\midrule
CUB & Glass blur & 69.56 & \textbf{5.96} & 52.31 & 36.82 \\
 & Noise + defocus & 13.67 & 12.79 & 11.83 & \textbf{1.73} \\
\bottomrule
\end{tabular*}
\end{table}

The fitted operator helps only when it lies close to the pool. With the library alone, glass blur fails on both datasets. $\text{Known}_c$ stays at the level of SimGCD (2.6 on \cifarhundred and 0.0 on \cub), and on \cifarhundred the matcher picks brightness at severity~1, which is nearly the identity. On \cub glass blur, the parametric fit is 5.96 floor units away and its displacement stage switches on, as glass blur's does. It lifts $\text{Known}_c$ to 47.8 and All accuracy to 8.8 points above SimGCD, at a cost of 3.9 points of clean Known accuracy against the library run. On \cifarhundred glass blur, the best fit is 15.0 units away, and the corrupted images collapse into 29 of the 100 clusters, as with a wrong library operator. On the two-operator pool, the free-form $k = 2$ fit reaches 0.15 and 1.73 floor units, which no single library operator does, and adds 7.3 and 5.4 points of All accuracy and 14.7 and 20.4 points of $\text{Known}_c$ over the library run. On the 52 in-library and $\rho = 0$ pools the rule never fires (the closest, \cub at $\rho = 0$, is blocked by criterion (A)), so no other result changes.

\begin{figure}[!htb]
  \centering
  \includegraphics[width=\textwidth]{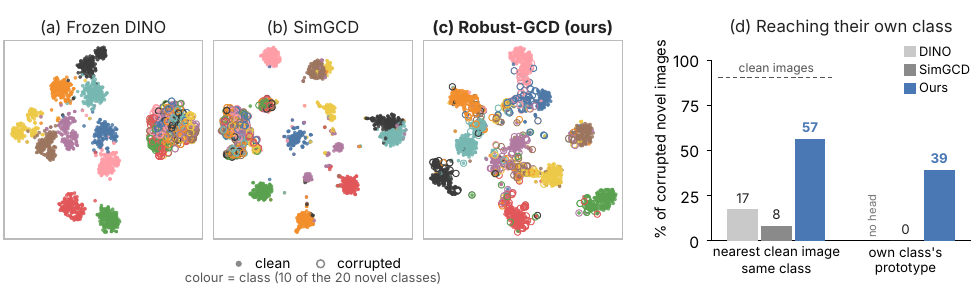}
  \caption{Where the corrupted novel images go. \cifarhundred, Gaussian noise,
  $\rho = 0.25$, final checkpoints (All: SimGCD 50.7, \method 69.7).
  (a)--(c)~t-SNE of 10 of the 20 novel classes, 200 unlabeled images per class;
  color is class, filled markers are clean images, hollow markers are corrupted.
  (d)~Measured in the full feature space over all 10{,}000 novel images: the share
  of corrupted images whose nearest clean image has the same class, and the share
  assigned to the prototype that holds most of their class's clean images. The
  dashed line is the first measure for clean images.}
  \label{fig:analysis}
\end{figure}

\textbf{Feature visualization.}
\label{sec:exp-vis}
\method moves corrupted novel images into the clusters of their own class (Figure~\ref{fig:analysis}). Figure~\ref{fig:analysis}(a)--(c) shows t-SNE~\citep{vandermaaten2008tsne} embeddings of the novel classes. In frozen DINO, the nearest neighbor of every corrupted novel image is another corrupted image (100\%, against a chance level of 25\%). SimGCD keeps this region and gives it prototypes of its own, and none of its corrupted novel images lands in the prototype that holds the clean images of its class (Figure~\ref{fig:analysis}(d)). Under \method, 39\% do, and 57\% have a clean image of their class as nearest neighbor (SimGCD: 8\%).

\FloatBarrier 

\section{Conclusion and Limitations}
\label{sec:conclusion}

In this paper, we formulate a more realistic scenario, \setting, which deals
with contamination in the unlabeled pool, and design a novel framework
\method to fight against this contamination. Experiments across five datasets
and several corruption families verify the effectiveness of the proposed
method. There are nevertheless some limitations. Our main experiments corrupt each pool with a single family from a fixed library. With a fitted operator, the method also extends to some corruptions outside the library (Section~\ref{sec:beyond}), but deciding when a fitted operator is close enough to trust is left to future work. We hope our work
can promote the research on robust category discovery in real-world
applications.

\section*{Acknowledgments}
We thank Hongfu Liu from Brandeis University for guidance throughout this project.

\bibliographystyle{unsrtnat}
\bibliography{references}

\newpage
\appendix
\counterwithin{table}{section}

\section{Per-Family Results}
\label{app:perfamily}

Tables~\ref{tab:pf-cifar10} to~\ref{tab:pf-aircraft} give the result of every
run behind the means of Table~\ref{tab:main}, one run per dataset and family,
with $\rho = 0.25$ and severity 5. Family names follow the implementation of
\citet{michaelis2019imagecorruptions}. Brightness and contrast, two DINO pretraining augmentations, are evaluated on \cub.

\begin{table}[!htbp]
  \centering
  \small
  \caption{Per-family results on \cifarten ($\rho = 0.25$, severity 5, one run per
  row). The clean and corrupted columns use corruption membership and are a
  diagnostic. Bold: the better of the two methods.}
  \label{tab:pf-cifar10}
  \begin{tabular*}{\textwidth}{@{\extracolsep{\fill}}ll ccc cc cc}
    \toprule
    & & \multicolumn{3}{c}{\textbf{All images}} & \multicolumn{2}{c}{\textbf{Clean}} & \multicolumn{2}{c}{\textbf{Corrupted}} \\
    \cmidrule(lr){3-5}\cmidrule(lr){6-7}\cmidrule(lr){8-9}
    \textbf{Family} & \textbf{Method} & \textbf{All} & \textbf{Known} & \textbf{Novel} & \textbf{Known} & \textbf{Novel} & \textbf{Known} & \textbf{Novel} \\
    \midrule
    Gaussian noise & SimGCD & 59.4 & 71.8 & 53.2 & \textbf{96.3} & 59.2 & 0.0 & 35.1 \\
     & \method & \textbf{92.3} & \textbf{92.7} & \textbf{92.2} & 95.9 & \textbf{97.8} & \textbf{83.2} & \textbf{75.1} \\
    \midrule
    Mean & SimGCD & 59.4 & 71.8 & 53.2 & \textbf{96.3} & 59.2 & 0.0 & 35.1 \\
     & \method & \textbf{92.3} & \textbf{92.7} & \textbf{92.2} & 95.9 & \textbf{97.8} & \textbf{83.2} & \textbf{75.1} \\
    \bottomrule
  \end{tabular*}
\end{table}

\begin{table}[!htbp]
  \centering
  \small
  \caption{Per-family results on \cifarhundred ($\rho = 0.25$, severity 5, one run per
  row). The clean and corrupted columns use corruption membership and are a
  diagnostic. Bold: the better of the two methods.}
  \label{tab:pf-cifar100}
  \begin{tabular*}{\textwidth}{@{\extracolsep{\fill}}ll ccc cc cc}
    \toprule
    & & \multicolumn{3}{c}{\textbf{All images}} & \multicolumn{2}{c}{\textbf{Clean}} & \multicolumn{2}{c}{\textbf{Corrupted}} \\
    \cmidrule(lr){3-5}\cmidrule(lr){6-7}\cmidrule(lr){8-9}
    \textbf{Family} & \textbf{Method} & \textbf{All} & \textbf{Known} & \textbf{Novel} & \textbf{Known} & \textbf{Novel} & \textbf{Known} & \textbf{Novel} \\
    \midrule
    Fog & SimGCD & 51.1 & 57.0 & 39.4 & 75.7 & 49.1 & 1.5 & 9.5 \\
     & \method & \textbf{68.3} & \textbf{72.7} & \textbf{59.4} & \textbf{80.5} & \textbf{68.1} & \textbf{49.6} & \textbf{32.9} \\
    \addlinespace[2pt]
    Frost & SimGCD & 50.7 & 57.0 & 38.2 & 75.7 & 48.4 & 1.3 & 6.9 \\
     & \method & \textbf{71.0} & \textbf{75.6} & \textbf{61.9} & \textbf{81.5} & \textbf{68.4} & \textbf{58.1} & \textbf{41.7} \\
    \addlinespace[2pt]
    Gaussian noise & SimGCD & 50.7 & 56.7 & 38.6 & 74.9 & 48.5 & 2.6 & 8.2 \\
     & \method & \textbf{69.7} & \textbf{75.3} & \textbf{58.7} & \textbf{81.0} & \textbf{65.8} & \textbf{58.5} & \textbf{36.8} \\
    \addlinespace[2pt]
    Impulse noise & SimGCD & 51.6 & 56.4 & 42.1 & 74.7 & 52.2 & 2.2 & 11.2 \\
     & \method & \textbf{73.4} & \textbf{76.4} & \textbf{67.5} & \textbf{81.7} & \textbf{75.2} & \textbf{60.7} & \textbf{43.9} \\
    \addlinespace[2pt]
    Shot noise & SimGCD & 50.6 & 58.8 & 34.2 & 78.1 & 41.1 & 1.5 & 12.9 \\
     & \method & \textbf{72.2} & \textbf{75.7} & \textbf{65.2} & \textbf{81.4} & \textbf{73.2} & \textbf{58.7} & \textbf{40.6} \\
    \addlinespace[2pt]
    Snow & SimGCD & 50.6 & 55.7 & 40.4 & 74.0 & 51.8 & 1.6 & 5.4 \\
     & \method & \textbf{69.6} & \textbf{77.4} & \textbf{54.0} & \textbf{83.0} & \textbf{59.4} & \textbf{60.8} & \textbf{37.4} \\
    \addlinespace[2pt]
    Spatter & SimGCD & 51.6 & 56.8 & 41.1 & 75.3 & 51.8 & 2.0 & 8.4 \\
     & \method & \textbf{71.1} & \textbf{78.8} & \textbf{55.7} & \textbf{82.8} & \textbf{59.9} & \textbf{66.7} & \textbf{42.9} \\
    \addlinespace[2pt]
    Speckle noise & SimGCD & 50.7 & 59.3 & 33.5 & 78.8 & 40.8 & 1.7 & 11.0 \\
     & \method & \textbf{69.7} & \textbf{77.5} & \textbf{54.0} & \textbf{82.4} & \textbf{57.6} & \textbf{63.1} & \textbf{43.1} \\
    \addlinespace[2pt]
    Zoom blur & SimGCD & 58.3 & 63.7 & 47.4 & 78.1 & 52.7 & 21.0 & 31.2 \\
     & \method & \textbf{79.7} & \textbf{80.5} & \textbf{78.1} & \textbf{83.0} & \textbf{80.7} & \textbf{73.2} & \textbf{70.4} \\
    \midrule
    Mean & SimGCD & 51.8 & 57.9 & 39.4 & 76.1 & 48.5 & 3.9 & 11.6 \\
     & \method & \textbf{71.6} & \textbf{76.7} & \textbf{61.6} & \textbf{81.9} & \textbf{67.6} & \textbf{61.0} & \textbf{43.3} \\
    \bottomrule
  \end{tabular*}
\end{table}

\begin{table}[!htbp]
  \centering
  \small
  \caption{Per-family results on \cub ($\rho = 0.25$, severity 5, one run per
  row). The clean and corrupted columns use corruption membership and are a
  diagnostic. Bold: the better of the two methods.}
  \label{tab:pf-cub}
  \begin{tabular*}{\textwidth}{@{\extracolsep{\fill}}ll ccc cc cc}
    \toprule
    & & \multicolumn{3}{c}{\textbf{All images}} & \multicolumn{2}{c}{\textbf{Clean}} & \multicolumn{2}{c}{\textbf{Corrupted}} \\
    \cmidrule(lr){3-5}\cmidrule(lr){6-7}\cmidrule(lr){8-9}
    \textbf{Family} & \textbf{Method} & \textbf{All} & \textbf{Known} & \textbf{Novel} & \textbf{Known} & \textbf{Novel} & \textbf{Known} & \textbf{Novel} \\
    \midrule
    Brightness & SimGCD & 44.6 & 49.2 & 42.3 & \textbf{64.0} & 48.4 & 5.3 & 23.8 \\
     & \method & \textbf{57.9} & \textbf{62.0} & \textbf{55.8} & 62.7 & \textbf{57.4} & \textbf{60.2} & \textbf{50.9} \\
    \addlinespace[2pt]
    Contrast & SimGCD & 41.7 & 49.7 & 37.7 & 65.4 & 47.5 & 2.9 & 8.4 \\
     & \method & \textbf{55.8} & \textbf{64.8} & \textbf{51.2} & \textbf{66.5} & \textbf{55.7} & \textbf{60.0} & \textbf{37.8} \\
    \addlinespace[2pt]
    Fog & SimGCD & 42.4 & 50.1 & 38.6 & 65.0 & 44.7 & 5.8 & 20.2 \\
     & \method & \textbf{58.4} & \textbf{65.7} & \textbf{54.8} & \textbf{65.4} & \textbf{56.5} & \textbf{66.6} & \textbf{49.5} \\
    \addlinespace[2pt]
    Frost & SimGCD & 41.4 & 51.3 & 36.4 & \textbf{67.6} & 45.4 & 2.9 & 9.2 \\
     & \method & \textbf{53.3} & \textbf{61.3} & \textbf{49.2} & \textbf{67.6} & \textbf{54.2} & \textbf{42.7} & \textbf{34.1} \\
    \addlinespace[2pt]
    Gaussian noise & SimGCD & 40.7 & 47.6 & 37.2 & \textbf{63.5} & 46.4 & 0.5 & 9.5 \\
     & \method & \textbf{51.3} & \textbf{59.2} & \textbf{47.4} & \textbf{63.5} & \textbf{53.4} & \textbf{46.4} & \textbf{29.3} \\
    \addlinespace[2pt]
    Impulse noise & SimGCD & 39.8 & 47.4 & 36.0 & 62.6 & 45.3 & 2.1 & 8.0 \\
     & \method & \textbf{54.2} & \textbf{61.7} & \textbf{50.5} & \textbf{65.4} & \textbf{55.2} & \textbf{50.7} & \textbf{36.5} \\
    \addlinespace[2pt]
    Shot noise & SimGCD & 41.6 & 50.0 & 37.4 & 66.3 & 46.9 & 1.3 & 8.8 \\
     & \method & \textbf{54.7} & \textbf{62.8} & \textbf{50.6} & \textbf{66.6} & \textbf{55.6} & \textbf{51.7} & \textbf{35.5} \\
    \addlinespace[2pt]
    Snow & SimGCD & 40.7 & 48.8 & 36.7 & 64.0 & 45.6 & 3.7 & 9.6 \\
     & \method & \textbf{56.7} & \textbf{62.8} & \textbf{53.7} & \textbf{65.1} & \textbf{58.4} & \textbf{56.0} & \textbf{39.5} \\
    \addlinespace[2pt]
    Spatter & SimGCD & 40.9 & 50.8 & 35.9 & 67.3 & 43.4 & 1.9 & 13.4 \\
     & \method & \textbf{52.2} & \textbf{60.0} & \textbf{48.2} & \textbf{68.7} & \textbf{54.1} & \textbf{34.0} & \textbf{30.7} \\
    \addlinespace[2pt]
    Speckle noise & SimGCD & 41.2 & 48.8 & 37.5 & 64.1 & 47.4 & 3.2 & 7.6 \\
     & \method & \textbf{58.3} & \textbf{64.0} & \textbf{55.4} & \textbf{64.6} & \textbf{59.6} & \textbf{62.3} & \textbf{42.6} \\
    \addlinespace[2pt]
    Zoom blur & SimGCD & 43.0 & 51.1 & 39.0 & \textbf{66.9} & 48.2 & 4.0 & 11.2 \\
     & \method & \textbf{54.3} & \textbf{62.6} & \textbf{50.2} & 63.8 & \textbf{54.5} & \textbf{58.9} & \textbf{37.4} \\
    \midrule
    Mean & SimGCD & 41.6 & 49.5 & 37.7 & 65.2 & 46.3 & 3.1 & 11.8 \\
     & \method & \textbf{55.2} & \textbf{62.4} & \textbf{51.5} & \textbf{65.4} & \textbf{55.9} & \textbf{53.6} & \textbf{38.5} \\
    \bottomrule
  \end{tabular*}
\end{table}

\begin{table}[!htbp]
  \centering
  \small
  \caption{Per-family results on \cars ($\rho = 0.25$, severity 5, one run per
  row). The clean and corrupted columns use corruption membership and are a
  diagnostic. Bold: the better of the two methods.}
  \label{tab:pf-cars}
  \begin{tabular*}{\textwidth}{@{\extracolsep{\fill}}ll ccc cc cc}
    \toprule
    & & \multicolumn{3}{c}{\textbf{All images}} & \multicolumn{2}{c}{\textbf{Clean}} & \multicolumn{2}{c}{\textbf{Corrupted}} \\
    \cmidrule(lr){3-5}\cmidrule(lr){6-7}\cmidrule(lr){8-9}
    \textbf{Family} & \textbf{Method} & \textbf{All} & \textbf{Known} & \textbf{Novel} & \textbf{Known} & \textbf{Novel} & \textbf{Known} & \textbf{Novel} \\
    \midrule
    Fog & SimGCD & 35.4 & 50.0 & 28.3 & 67.9 & 36.2 & 0.0 & 4.0 \\
     & \method & \textbf{53.3} & \textbf{70.6} & \textbf{44.9} & \textbf{71.3} & \textbf{46.1} & \textbf{68.4} & \textbf{41.2} \\
    \addlinespace[2pt]
    Frost & SimGCD & 33.2 & 48.3 & 25.9 & 65.6 & 33.7 & 0.0 & 1.7 \\
     & \method & \textbf{53.0} & \textbf{70.8} & \textbf{44.4} & \textbf{73.4} & \textbf{47.5} & \textbf{63.3} & \textbf{34.6} \\
    \addlinespace[2pt]
    Gaussian noise & SimGCD & 34.4 & 45.0 & 29.3 & 61.0 & 38.3 & 0.0 & 1.7 \\
     & \method & \textbf{45.4} & \textbf{67.3} & \textbf{34.9} & \textbf{72.5} & \textbf{38.6} & \textbf{52.5} & \textbf{23.3} \\
    \addlinespace[2pt]
    Impulse noise & SimGCD & 32.2 & 44.3 & 26.3 & 60.1 & 34.3 & 0.0 & 1.7 \\
     & \method & \textbf{47.1} & \textbf{68.2} & \textbf{36.9} & \textbf{72.1} & \textbf{40.1} & \textbf{57.0} & \textbf{27.2} \\
    \addlinespace[2pt]
    Shot noise & SimGCD & 34.7 & 47.9 & 28.3 & 65.0 & 36.9 & 0.0 & 1.7 \\
     & \method & \textbf{47.5} & \textbf{68.1} & \textbf{37.6} & \textbf{71.8} & \textbf{40.2} & \textbf{57.6} & \textbf{29.5} \\
    \addlinespace[2pt]
    Snow & SimGCD & 33.8 & 46.8 & 27.5 & 63.5 & 35.8 & 0.0 & 1.7 \\
     & \method & \textbf{46.6} & \textbf{68.0} & \textbf{36.2} & \textbf{69.9} & \textbf{38.4} & \textbf{62.5} & \textbf{29.5} \\
    \addlinespace[2pt]
    Spatter & SimGCD & 34.2 & 46.6 & 28.3 & 63.2 & 36.8 & 0.0 & 1.7 \\
     & \method & \textbf{49.1} & \textbf{73.2} & \textbf{37.5} & \textbf{75.1} & \textbf{39.2} & \textbf{67.9} & \textbf{32.3} \\
    \addlinespace[2pt]
    Speckle noise & SimGCD & 31.3 & 44.4 & 24.9 & 59.7 & 32.4 & 1.7 & 1.6 \\
     & \method & \textbf{49.3} & \textbf{69.5} & \textbf{39.5} & \textbf{71.3} & \textbf{41.2} & \textbf{64.5} & \textbf{34.3} \\
    \addlinespace[2pt]
    Zoom blur & SimGCD & 31.8 & 44.2 & 25.7 & 60.0 & 33.5 & 0.0 & 1.7 \\
     & \method & \textbf{50.5} & \textbf{70.6} & \textbf{40.8} & \textbf{73.4} & \textbf{43.6} & \textbf{62.9} & \textbf{32.3} \\
    \midrule
    Mean & SimGCD & 33.4 & 46.4 & 27.2 & 62.9 & 35.3 & 0.2 & 1.9 \\
     & \method & \textbf{49.1} & \textbf{69.6} & \textbf{39.2} & \textbf{72.3} & \textbf{41.7} & \textbf{61.8} & \textbf{31.6} \\
    \bottomrule
  \end{tabular*}
\end{table}

\begin{table}[!htbp]
  \centering
  \small
  \caption{Per-family results on \aircraft ($\rho = 0.25$, severity 5, one run per
  row). The clean and corrupted columns use corruption membership and are a
  diagnostic. Bold: the better of the two methods.}
  \label{tab:pf-aircraft}
  \begin{tabular*}{\textwidth}{@{\extracolsep{\fill}}ll ccc cc cc}
    \toprule
    & & \multicolumn{3}{c}{\textbf{All images}} & \multicolumn{2}{c}{\textbf{Clean}} & \multicolumn{2}{c}{\textbf{Corrupted}} \\
    \cmidrule(lr){3-5}\cmidrule(lr){6-7}\cmidrule(lr){8-9}
    \textbf{Family} & \textbf{Method} & \textbf{All} & \textbf{Known} & \textbf{Novel} & \textbf{Known} & \textbf{Novel} & \textbf{Known} & \textbf{Novel} \\
    \midrule
    Fog & SimGCD & 38.5 & 41.1 & 37.1 & 53.1 & 44.4 & 4.6 & 15.4 \\
     & \method & \textbf{53.2} & \textbf{61.6} & \textbf{48.9} & \textbf{64.5} & \textbf{49.9} & \textbf{52.9} & \textbf{46.2} \\
    \addlinespace[2pt]
    Frost & SimGCD & 40.7 & 44.4 & 38.8 & 59.1 & \textbf{49.9} & 0.0 & 5.6 \\
     & \method & \textbf{48.3} & \textbf{60.7} & \textbf{42.0} & \textbf{63.0} & 44.5 & \textbf{53.6} & \textbf{34.7} \\
    \addlinespace[2pt]
    Gaussian noise & SimGCD & 38.2 & 44.4 & \textbf{35.0} & \textbf{59.1} & \textbf{45.1} & 0.0 & 5.1 \\
     & \method & \textbf{41.3} & \textbf{55.8} & 34.1 & 59.0 & 36.0 & \textbf{46.1} & \textbf{28.3} \\
    \addlinespace[2pt]
    Impulse noise & SimGCD & 41.6 & 47.7 & \textbf{38.6} & \textbf{63.5} & \textbf{49.5} & 0.0 & 6.0 \\
     & \method & \textbf{42.9} & \textbf{57.7} & 35.5 & 60.6 & 38.7 & \textbf{49.0} & \textbf{25.8} \\
    \addlinespace[2pt]
    Shot noise & SimGCD & \textbf{41.2} & 46.8 & \textbf{38.5} & \textbf{62.2} & \textbf{49.4} & 0.0 & 5.9 \\
     & \method & \textbf{41.2} & \textbf{52.4} & 35.6 & 55.6 & 38.6 & \textbf{42.8} & \textbf{26.4} \\
    \addlinespace[2pt]
    Snow & SimGCD & 38.5 & 45.4 & 35.0 & 60.1 & \textbf{44.6} & 1.0 & 6.2 \\
     & \method & \textbf{47.3} & \textbf{59.4} & \textbf{41.3} & \textbf{61.4} & 43.4 & \textbf{53.4} & \textbf{34.9} \\
    \addlinespace[2pt]
    Spatter & SimGCD & 38.6 & 44.7 & 35.6 & 59.4 & \textbf{46.5} & 0.0 & 3.4 \\
     & \method & \textbf{46.9} & \textbf{58.7} & \textbf{40.9} & \textbf{61.4} & 42.9 & \textbf{50.5} & \textbf{34.9} \\
    \addlinespace[2pt]
    Speckle noise & SimGCD & 42.5 & 47.3 & 40.2 & \textbf{62.9} & \textbf{51.3} & 0.0 & 6.7 \\
     & \method & \textbf{46.7} & \textbf{56.9} & \textbf{41.6} & 58.9 & 43.9 & \textbf{51.0} & \textbf{34.7} \\
    \addlinespace[2pt]
    Zoom blur & SimGCD & 40.8 & 44.1 & 39.2 & 57.7 & \textbf{47.9} & 3.1 & 13.3 \\
     & \method & \textbf{49.4} & \textbf{59.7} & \textbf{44.3} & \textbf{61.6} & 45.9 & \textbf{54.1} & \textbf{39.6} \\
    \midrule
    Mean & SimGCD & 40.1 & 45.1 & 37.6 & 59.7 & \textbf{47.6} & 1.0 & 7.5 \\
     & \method & \textbf{46.4} & \textbf{58.1} & \textbf{40.5} & \textbf{60.7} & 42.6 & \textbf{50.4} & \textbf{33.9} \\
    \bottomrule
  \end{tabular*}
\end{table}

\clearpage % each appendix section starts on its own page, after the previous one's tables
\section{Corruption Estimation}
\label{app:estimation}

Tables~\ref{tab:estimation} and~\ref{tab:estimation-b} report, for every pool of Table~\ref{tab:main} and Figure~\ref{fig:rho}, what the corruption estimation of Section~\ref{sec:estimate} recovers, scored against the true corruption after the fact. None of these values is available to the method. $\hat\rho$ is Storey's estimate of $\rho$ (Eq.~\eqref{eq:storey}). Precision is the share of the target set $\mathcal{T}$ that is truly corrupted. The Stage One column is the candidate with the smallest $W_1^{\text{avg}}$, which is the operator Stage One alone would choose, and $(c^\ast, s^\ast)$ is the final choice after Stage Two. The margin is the gap between the best and second-best $\mathrm{MMD}^2$ in units of the floor $\mathrm{MMD}^2(\Dl, \mathcal{R})$.

The matcher names the true family on all 47 pools and the true family and severity on 41. Every miss is a severity one step below the truth (fog on three datasets, \cifarhundred zoom blur and \cub spatter), except \cub frost, matched at severity 2. The true pair is in the Stage One shortlist on all 47 pools. Stage One alone names the wrong family on three of the four zoom-blur pools, choosing motion blur, and Stage Two corrects all three. 

\begin{table}[p]
\centering
\small
\caption{Corruption estimation per pool (severity 5). $\hat\rho$: estimated contamination rate. Prec.: share of $\mathcal{T}$ that is corrupted (\%). Stage One: the operator with the smallest $W_1^{\text{avg}}$. $(c^\ast, s^\ast)$: the matched operator. Margin: gap between the two smallest $\mathrm{MMD}^2$ in floor units.}
\label{tab:estimation}
\begin{tabular*}{\textwidth}{@{\extracolsep{\fill}}ll ccc ll c}
\toprule
\textbf{Dataset} & \textbf{Family} & \textbf{$\rho$} & \textbf{$\hat\rho$} & \textbf{Prec.} & \textbf{Stage One} & \textbf{$(c^\ast, s^\ast)$} & \textbf{Margin} \\
\midrule
CIFAR-10 & Gaussian noise & 0.25 & 0.20 & 100 & Gaussian noise, 5 & Gaussian noise, 5 & 0.62 \\
\midrule
CIFAR-100 & Fog & 0.25 & 0.25 & 98 & Fog, 5 & Fog, 5 & 24.56 \\
  & Frost & 0.25 & 0.25 & 98 & Frost, 5 & Frost, 5 & 0.31 \\
  & Gaussian noise & 0.05 & 0.07 & 78 & Gaussian noise, 5 & Gaussian noise, 5 & 0.39 \\
  & Gaussian noise & 0.1 & 0.11 & 91 & Gaussian noise, 5 & Gaussian noise, 5 & 1.45 \\
  & Gaussian noise & 0.25 & 0.25 & 98 & Gaussian noise, 5 & Gaussian noise, 5 & 2.64 \\
  & Gaussian noise & 0.5 & 0.50 & 100 & Gaussian noise, 5 & Gaussian noise, 5 & 24.05 \\
  & Gaussian noise & 1 & 1.00 & 100 & Gaussian noise, 5 & Gaussian noise, 5 & 0.02 \\
  & Impulse noise & 0.25 & 0.25 & 98 & Impulse noise, 5 & Impulse noise, 5 & 2.70 \\
  & Shot noise & 0.25 & 0.25 & 98 & Shot noise, 5 & Shot noise, 5 & 1.59 \\
  & Snow & 0.25 & 0.25 & 98 & Snow, 5 & Snow, 5 & 30.66 \\
  & Spatter & 0.25 & 0.25 & 98 & Spatter, 5 & Spatter, 5 & 9.21 \\
  & Speckle noise & 0.25 & 0.25 & 98 & Speckle noise, 5 & Speckle noise, 5 & 1.71 \\
  & Zoom blur & 0.25 & 0.25 & 73 & Motion blur, 5 & Zoom blur, 4 & 0.70 \\
\bottomrule
\end{tabular*}
\end{table}

\begin{table}[p]
\centering
\small
\caption{Corruption estimation per pool, continued (severity~5). Columns as in Table~\ref{tab:estimation}.}
\label{tab:estimation-b}
\begin{tabular*}{\textwidth}{@{\extracolsep{\fill}}ll ccc ll c}
\toprule
\textbf{Dataset} & \textbf{Family} & \textbf{$\rho$} & \textbf{$\hat\rho$} & \textbf{Prec.} & \textbf{Stage One} & \textbf{$(c^\ast, s^\ast)$} & \textbf{Margin} \\
\midrule
CUB & Brightness & 0.25 & 0.21 & 83 & Brightness, 5 & Brightness, 5 & 0.19 \\
  & Contrast & 0.25 & 0.21 & 100 & Contrast, 5 & Contrast, 5 & 15.22 \\
  & Fog & 0.25 & 0.21 & 94 & Fog, 3 & Fog, 4 & 5.06 \\
  & Frost & 0.25 & 0.19 & 65 & Frost, 5 & Frost, 2 & 0.64 \\
  & Gaussian noise & 0.05 & 0.02 & 100 & Gaussian noise, 5 & Gaussian noise, 5 & 0.65 \\
  & Gaussian noise & 0.1 & 0.07 & 100 & Gaussian noise, 5 & Gaussian noise, 5 & 0.86 \\
  & Gaussian noise & 0.25 & 0.21 & 100 & Gaussian noise, 5 & Gaussian noise, 5 & 1.64 \\
  & Gaussian noise & 0.5 & 0.48 & 100 & Gaussian noise, 5 & Gaussian noise, 5 & 3.04 \\
  & Gaussian noise & 1 & 1.00 & 100 & Gaussian noise, 5 & Gaussian noise, 5 & 0.01 \\
  & Impulse noise & 0.25 & 0.21 & 100 & Impulse noise, 5 & Impulse noise, 5 & 1.66 \\
  & Shot noise & 0.25 & 0.21 & 100 & Shot noise, 5 & Shot noise, 5 & 3.26 \\
  & Snow & 0.25 & 0.21 & 84 & Snow, 5 & Snow, 5 & 7.44 \\
  & Spatter & 0.25 & 0.18 & 61 & Spatter, 5 & Spatter, 4 & 2.50 \\
  & Speckle noise & 0.25 & 0.21 & 100 & Speckle noise, 5 & Speckle noise, 5 & 5.60 \\
  & Zoom blur & 0.25 & 0.20 & 90 & Motion blur, 4 & Zoom blur, 5 & 0.03 \\
\midrule
Stanford Cars & Fog & 0.25 & 0.22 & 99 & Fog, 3 & Fog, 4 & 4.47 \\
  & Frost & 0.25 & 0.22 & 91 & Frost, 5 & Frost, 5 & 0.14 \\
  & Gaussian noise & 0.25 & 0.22 & 100 & Gaussian noise, 5 & Gaussian noise, 5 & 1.26 \\
  & Impulse noise & 0.25 & 0.22 & 100 & Impulse noise, 5 & Impulse noise, 5 & 1.12 \\
  & Shot noise & 0.25 & 0.22 & 100 & Shot noise, 5 & Shot noise, 5 & 2.28 \\
  & Snow & 0.25 & 0.22 & 92 & Snow, 5 & Snow, 5 & 5.14 \\
  & Spatter & 0.25 & 0.22 & 80 & Spatter, 5 & Spatter, 5 & 0.95 \\
  & Speckle noise & 0.25 & 0.22 & 100 & Speckle noise, 5 & Speckle noise, 5 & 1.50 \\
  & Zoom blur & 0.25 & 0.22 & 96 & Zoom blur, 5 & Zoom blur, 5 & 0.04 \\
\midrule
FGVC-Aircraft & Fog & 0.25 & 0.27 & 92 & Fog, 3 & Fog, 4 & 6.76 \\
  & Frost & 0.25 & 0.27 & 89 & Frost, 5 & Frost, 5 & 0.49 \\
  & Gaussian noise & 0.25 & 0.27 & 92 & Gaussian noise, 5 & Gaussian noise, 5 & 1.38 \\
  & Impulse noise & 0.25 & 0.27 & 92 & Impulse noise, 5 & Impulse noise, 5 & 1.73 \\
  & Shot noise & 0.25 & 0.27 & 92 & Shot noise, 5 & Shot noise, 5 & 2.25 \\
  & Snow & 0.25 & 0.27 & 92 & Snow, 5 & Snow, 5 & 6.44 \\
  & Spatter & 0.25 & 0.27 & 92 & Spatter, 5 & Spatter, 5 & 4.74 \\
  & Speckle noise & 0.25 & 0.27 & 92 & Speckle noise, 5 & Speckle noise, 5 & 1.88 \\
  & Zoom blur & 0.25 & 0.27 & 92 & Motion blur, 3 & Zoom blur, 5 & 0.34 \\
\bottomrule
\end{tabular*}
\end{table}

\clearpage
\section{Clean-Pool Results}
\label{app:numbers}

Table~\ref{tab:vanilla} reports SimGCD, HiLo and \method on uncontaminated pools ($\rho = 0$), the runs behind the clean-pool comparison of Section~\ref{sec:exp-main} and the $\rho = 0$ points of Figures~\ref{fig:contamination} and~\ref{fig:rho}.

\begin{table}[!htbp]
\centering
\small
\caption{Accuracy (\%) on clean pools ($\rho = 0$). Bold: best per dataset and column. Single run per cell.}
\label{tab:vanilla}
\begin{tabular*}{\textwidth}{@{\extracolsep{\fill}}ll ccc}
\toprule
\textbf{Dataset} & \textbf{Method} & \textbf{All} & \textbf{Known} & \textbf{Novel} \\
\midrule
CIFAR-10 & SimGCD & 97.0 & 94.2 & \textbf{98.4} \\
 & HiLo & 74.5 & 91.5 & 66.0 \\
 & \method & \textbf{97.4} & \textbf{95.8} & 98.2 \\
\midrule
CIFAR-100 & SimGCD & \textbf{82.2} & 81.9 & \textbf{82.9} \\
 & HiLo & 77.6 & 80.4 & 72.1 \\
 & \method & 80.2 & \textbf{82.1} & 76.4 \\
\midrule
CUB & SimGCD & 61.1 & 64.4 & 59.5 \\
 & HiLo & 59.0 & \textbf{69.5} & 53.7 \\
 & \method & \textbf{63.0} & 67.7 & \textbf{60.7} \\
\midrule
Stanford Cars & SimGCD & 49.7 & 67.8 & \textbf{41.0} \\
 & HiLo & 49.6 & 67.5 & \textbf{41.0} \\
 & \method & \textbf{50.9} & \textbf{71.6} & 40.9 \\
\midrule
FGVC-Aircraft & SimGCD & 54.6 & 61.8 & \textbf{51.0} \\
 & HiLo & 49.0 & 45.3 & 50.8 \\
 & \method & \textbf{55.0} & \textbf{63.2} & 50.9 \\
\bottomrule
\end{tabular*}
\end{table}

\end{document}